\documentclass[conference]{IEEEtran}
\IEEEoverridecommandlockouts
\usepackage{cite}
\usepackage{amsmath,amssymb,amsfonts}
\usepackage{algorithm}
\usepackage{algorithmic}
\usepackage{graphicx}
\usepackage{textcomp}
\usepackage{xcolor}
\usepackage{epstopdf}
\usepackage{subcaption}
\newtheorem{theorem}{Theorem}
\newtheorem{definition}{Definition}
\newtheorem{lemma}{Lemma}
\newtheorem{proof}{Proof}
\allowdisplaybreaks
\def\BibTeX{{\rm B\kern-.05em{\sc i\kern-.025em b}\kern-.08em
    T\kern-.1667em\lower.7ex\hbox{E}\kern-.125emX}}
\begin{document}

\title{Soft policy optimization using dual-track advantage estimator}
\author{\IEEEauthorblockN{1\textsuperscript{st} Yubo Huang}
\IEEEauthorblockA{\textit{Department of Automation} \\
\textit{Shanghai Jiao Tong University}\\
Shanghai, China \\
huangyubo@mail.nwpu.edu.cn}
\\
\IEEEauthorblockN{4\textsuperscript{th} Zhiwei Zhuang}
\IEEEauthorblockA{\textit{Department of Automation} \\
\textit{Shanghai Jiao Tong University}\\
Shanghai, China \\
zzw1993@sjtu.edu.cn }
\and
\IEEEauthorblockN{2\textsuperscript{nd} Xuechun Wang}
\IEEEauthorblockA{\textit{Department of Automation} \\
\textit{Shanghai Jiao Tong University}\\
Shanghai, China \\
xuechun$\_$wang@sjtu.edu.cn}
\\
\IEEEauthorblockN{5\textsuperscript{th} Weidong Zhang\textsuperscript{$\star$}\thanks{$^\star$ Corresponding author.}}
\IEEEauthorblockA{\textit{Department of Automation} \\
\textit{Shanghai Jiao Tong University}\\
Shanghai, China \\
wdzhang@sjtu.edu.cn}
\and
\IEEEauthorblockN{3\textsuperscript{rd} Luobao Zou}
\IEEEauthorblockA{\textit{Department of Automation} \\
\textit{Shanghai Jiao Tong University}\\
Shanghai, China \\
leiling@sjtu.edu.cn}
}

\maketitle

\begin{abstract}
In reinforcement learning (RL), we always expect the agent to explore as many states as possible in the initial stage of training and exploit the explored information in the subsequent stage to discover the most returnable trajectory. Based on this principle, in this paper, we soften the proximal policy optimization by introducing the entropy and dynamically setting the temperature coefficient to balance the opportunity of exploration and exploitation. While maximizing the expected reward, the agent will also seek other trajectories to avoid the local optimal policy. Nevertheless, the increase of randomness induced by entropy will reduce the train speed in the early stage. Integrating the temporal-difference (TD) method and the general advantage estimator (GAE), we propose the dual-track advantage estimator (DTAE) to accelerate the convergence of value functions and further enhance the performance of the algorithm. Compared with other on-policy RL algorithms on the Mujoco environment, the proposed method not only significantly speeds up the training but also achieves the most advanced results in cumulative return.
\end{abstract}


\section{Introduction}
Deep reinforcement learning algorithms, which combine the classical RL framework and the high-capacity function approximators (i.e. neural networks) have achieved tremendous advanced results in complicate decision-making tasks such as robotic control~\cite{akkaya2019solving}, recommendation systems~\cite{ie2019reinforcement} and game playing~\cite{schrittwieser2019mastering}, etc. We can divide them into two categories: model-based or model-free RL. In model-based RL, we should learn not only the policy but the model in the optimization. Therefore, model-based RL allows deeper cognition of the environment but it is of storage and time cost since the mapping space from state-action-reward to its next state is extremely huge. The model error as well as the value function error is introduced in the learning. Considering that it is difficult to construct a sufficiently accurate environment in challenging robot control tasks, we focus on the model-free RL to train the agents in this paper.

On-policy learning and off-policy learning are two branches of model-free RL. On-policy RL algorithms require collecting new samples which are generated by the current policy to optimize the policy function at each gradient step. TRPO~\cite{schulman2015trust} is one of the representative methods of on-policy RL but it is relatively complicated (second-order optimization) to compute and is incompatible with parameter sharing structure such as between the policy and value function or architectures that include noise~\cite{schulman2017proximal}. PPO~\cite{schulman2017proximal} which uses the clipped trick and ACKTR~\cite{wu2017scalable} which uses the Kronecker-factored approximate curvature are proposed to reduce the computational complexity and expand the application scope of trust region methods. Although many studies show the high effectiveness of on-policy algorithms~\cite{zou2019separated,ahmed2018understanding,liu2019policy}, they are still criticized for sample inefficient because of the large demand for new samples at each batch. Off-policy RL algorithms use the experience buffer to reuse the past samples and thereupon are data efficient. The main contenders are Q-function based methods~\cite{mnih2013playing,wang2015dueling} and actor-critic~\cite{mnih2016asynchronous}. For example, Schaul et al. proposed the prioritized experience replay to schedule samples and further speed up the optimization of the value function~\cite{schaul2015prioritized}. Hasselt et al. proposed double Q-learning~\cite{van2016deep} and Fujimoto et al. proposed TD3~\cite{fujimoto2018addressing} to solve the overestimation problem in off-policy RL. Lillicrap et al. combined actor-critic and deterministic policy gradient to learn competitive policies for tasks with the continuous action space~\cite{lillicrap2015continuous}, which are difficult for value based off-policy RL algorithms. Haarnoja introduced the maximum entropy framework to actor-critic to increase the exploration of the agent~\cite{haarnoja2018soft}. Nevertheless, the algorithms which combine the off-policy learning RL and deep neural networks present challenges in terms of stability and convergence, especially for high-dimensional continuous control tasks~\cite{wang2016sample,bhatnagar2009convergent}.

From above analysis, there are two deficiencies that limit the application of model-free RL: (1) the algorithms are sample inefficiency and require tons of samples to optimize the value and policy function; (2) some algorithms are difficult to converge and are sensitive to hyper-parameters or time seeds. Therefore, in this paper, we aim to design a reliable and efficient on-policy RL algorithm for challenging continuous robotic control tasks. First, we introduce the entropy term to the objective to balance the opportunity of exploration and exploitation in RL (soft policy optimization). By dynamically setting the temperature coefficient, the agent will explore more states in the initial stage of training and subsequently exploit the explored policy to find more returnable trajectories. Nevertheless, the cumulative rate of return in the early stage will also be weakened meanwhile since the agent tends to adopt a more stochastic action rather than the greedy deterministic action. To tackle this problem, we present the concept of shadow value function and shadow policy function. We use the TD method to update the shadow value function and derive the TD advantage estimator (TDAE). TD methods have the faster convergence speed but the value functions may be unstable during the optimization. On the contrary, GAE is more cautious when updating their parameter vectors. Integrating TDAE and GAE, we propose the dual-track advantage estimator to accelerate the optimization of value function and indirectly improve the sample utilization efficiency. Theoretically, we have strictly proved that the soft policy optimization can improve the policy in each iteration. Results show the proposed algorithm called SPOD can not only significantly speed up the training in the early stage but also performs excellent in accumulating return.

\section{Preliminaries}
\subsection{The basic notation of reinforcement learning}
We can standardize the interacting of an agent with its environment as a Markov decision process (MDP) in reinforcement learning. At each discrete time step $t$, the agent observes a state $s_t \in \mathcal{S}$ of the environment, and samples an action $a \in \mathcal{A}$ from the policy $\pi : \mathcal{S} \rightarrow \mathcal{A}$. Then, the environment will feedback a reward $r \in \mathcal{R}$ and jump to the next state $s_{t+1}$ based on the transition probability distribution $P: \mathcal{S} \times \mathcal{A} \rightarrow \mathcal{S}$. The objective is to optimize the appropriate policy $\pi_{\theta}$ ($\theta$ is the parameters of the neural network) which maximizes the expected return:
\begin{equation}\label{return}
  J(\pi) = \mathbb{E}_{(s_t, a_t)\sim \rho(s_0), \pi, p}\{\sum_{t=0}^{\infty} \gamma^t r(s_t, a_t)\}
\end{equation}
where $\rho$ is the distribution of the initial state $s_0$.

Based on the policy $\pi$, the state value function $V_{\pi}(s_t)$, the state-action value function $Q_{\pi}(s_t, a_t)$ , and the advantage function $A_{\pi}(s_t, a_t)$ can be defined as:
\begin{equation}\label{value function}
  \begin{split}
       & V_{\pi}(s_t) = \mathbb{E}_{a_t, s_t,... \sim \pi, p} \{\sum_{l=0}^{\infty} \gamma^l r(s_{t+l})\} \\
       &  Q_{\pi}(s_t, a_t) = \mathbb{E}_{s_{t+1},a_{t+1},... \sim p, \pi} \{\sum_{l=0}^{\infty} \gamma^l r(s_{t+l}, a_{t+l})\}  \\
       & A_{\pi}(s_t, a_t) = Q(s_t, a_t) - V(s_t)
  \end{split}
\end{equation}

\subsection{Trust region and proximal policy optimization}
Kakade~\cite{kakade2002approximately} and Schulman~\cite{schulman2015trust} derived that the sufficient condition to increase the policy performance in a policy update $\pi \rightarrow \tilde{\pi} $ is: $\sum_a \tilde{\pi}(a|s) A_{\pi}(s,a) \geq 0$. Based on this principle, they developed the trust region policy optimization (TRPO) method which its objective function is:

\begin{equation}\label{trpo}
  \begin{split}
      & \max_{\theta} \quad \mathbb{\hat{E}}_t \left [\frac{\pi_{\theta}(a_t|s_t)}{\pi_{\theta_{old}(a_t|s_t)}} \hat{A}_t \right] \\
      & s.t. \quad \mathbb{\hat{E}}_t \left [D_{KL}(\pi_{\theta_{old}}(\cdot | s) || \pi_{\theta}(\cdot | s)) \right] \leq \delta
  \end{split}
\end{equation}

where $\hat{A}_t = A_{\pi_{\theta_{old}}}(s_t,a_t)$, $\pi_{\theta_{old}}$ is the policy before updating and $\delta$ is the trust region used to ensure the convergence of TRPO. They introduced the Lagrangian multiplier and conjugate gradient methods to solve the optimization problem but the time cost is expensive. Meanwhile, $\delta$ is difficult to determine in different environments. Therefore, Schulman~\cite{schulman2017proximal} proposed the surrogate objective (PPO) with the clipped probability ratio $r_t = \pi_\theta(a_t|s_t) / \pi_{\theta_{old}}(a_t|s_t)$ to remove the restricted condition in TRPO objective~(Eq.~\ref{trpo}):

\begin{equation}\label{ppo}
 \max_{\theta} \quad \mathbb{\hat{E}}_t \left [\min \left( r_t \hat{A}_t, clip(r_t, 1-\epsilon, 1+\epsilon) \hat{A}_t\right) \right]
\end{equation}
where $\epsilon$ is the clip margin which enables the final objective is a lower bound on the unclipped objective.
\section{Temporal-Difference based advantage prediction}
Before introducing the proposed method, we define that $\theta$ and $\phi$ are the parameter vectors of the policy network $\pi$ and the value network $V$, respectively. $\pi_{\theta_{k-1}}$ (the policy in $k-1$th iteration) is the shadow policy of $\pi_{\theta_{k}}$ and $V_{\phi_{k-1}}$ is the shadow value function of $V_{\phi_{k}}$.
\subsection{The generalized advantage estimator (GAE)}
Normally, we can write the advantage value $A(s_t, a_t)$ with the form of temporal-difference (TD) error $\delta_t$:
\begin{equation}\label{delta}
  \begin{split}
   A(s_t, a_t) & = Q_{\phi}(s_t, a_t) - V_{\phi}(s_t) = G_t - V_{\phi}(s_t) \\
               & = r_{t+1} + \gamma V_{\phi}(s_{t+1}) - V_{\phi}(s_t) = \delta_t
  \end{split}
\end{equation}
where $G_t$ is the target for the Monte Carlo update at time $t$, $\gamma$ is the discount factor and $\delta_t$ is the TD error in $V_{\phi}(s_t)$. In fact, the above update target $G_t$ is just one-step cumulative reward and we can extend it into multi-steps cumulative reward in an episode:
\begin{equation}\label{nreturen}
  G_{t:t+n} = r_{t+1} + \gamma r_{t+2} + ... + \gamma^{n-1} r_{t+n} + \gamma^n V_{\phi}(s_{t+n})
\end{equation}
Then, $G_{t:t+k}$ can be written as the first $n$ TD errors plus the estimated value of $s_t$~\cite{sutton2018reinforcement}:
\begin{equation}\label{nTDreturen}
  G_{t:t+n} =  V_{\phi}(s_t) + \sum_{k=0}^{n} \gamma^{k}\delta_{s_{t+k}}
\end{equation}
%\begin{equation}\label{nTDreturen}
%\begin{split}
%  G_{t:t+1} = & r_{t+1} + \gamma V_{\phi}(s_{t+1}) \\
%            = & V_{\phi}(s_t) + r_{t+1} + \gamma V_{\phi}(s_{t+1}) - V_{\phi}(s_t) \\
%            = & V_{\phi}(s_t) + \delta_{s_t} \\
%  G_{t:t+2} = & r_{t+1} + \gamma r_{t+2} + \gamma^2 V_{\phi}(s_{t+2}) \\
%            = & V_{\phi}(s_t) + r_{t+1} + \gamma V_{\phi}(s_{t+1}) - V_{\phi}(s_t)  \\
%              & + \gamma (r_{t+2} + \gamma V_{\phi}(s_{t+2}) - V_{\phi}(s_{t+1})) \\
%            = & V_{\phi}(s_t) + \delta_{s_t} + \gamma \delta_{s_{t+1}} \\
%  \vdots \qquad   & \quad \vdots \qquad \quad  \vdots \\
%  G_{t:t+n} = & V_{\phi}(s_t) + \sum_{k=0}^{n} \gamma^{k}\delta_{s_{t+k}}
%\end{split}
%\end{equation}

Sutton~\cite{sutton2018reinforcement} defined the $\lambda$ return as:

\begin{equation}\label{lambdareturn}
  \begin{split}
     G_t^\lambda & = (1-\lambda) \sum_{n=1}^{\infty}\lambda^{n-1} G_{t:t+n}\\
      & = V_{\phi}(s_t) + \sum_{k=0}^{\infty} (\gamma \lambda)^k \delta_{t+k}
  \end{split}
\end{equation}
%\begin{equation}\label{lambdareturn}
%  \begin{split}
%     G_t^\lambda & = (1-\lambda) \sum_{n=1}^{\infty}\lambda^{n-1} G_{t:t+n}\\
%      & = (1-\lambda) \sum_{n=1}^{\infty}\lambda^{n-1}(V_{\phi}(s_t) + \sum_{k=0}^{n} \gamma^{k}\delta_{s_{t+k}})\\
%      & = V_{\phi}(s_t) + (1-\lambda)[\delta_{s_{t}} + \lambda(\delta_{s_t} + \gamma\delta_{s_{t+1}}) + \cdots] \\
%      & = V_{\phi}(s_t) + (1-\lambda)[\delta_{s_t}(1 + \lambda + \lambda^2 + \cdots) \\
%      & \quad + \gamma \delta_{s_{t+1}}(\lambda + \lambda^2 + + \lambda^3 + \cdots) \\
%      & \quad + \gamma^2 \delta_{s_{t+2}}(\lambda^2 + \lambda^3 + \lambda^4 + \cdots) + \cdots]\\
%      & = V_{\phi}(s_t) + (1-\lambda)[\delta_{s_t}(\frac{1}{1-\lambda}) + \delta_{s_{t+1}}(\frac{\lambda}{1-\lambda})+\cdots ]\\
%      & = V_{\phi}(s_t) + \sum_{k=0}^{\infty} (\gamma \lambda)^k \delta_{t+k} \\
%  \end{split}
%\end{equation}
Based on the $\lambda$ return, Schulman~\cite{schulman2015high} proposed the general advantage function (GAE):
\begin{equation}\label{gae}
    A^{G} = G_{t}^{\lambda} - V_{\phi}(s_t) = \sum_{k=0}^{\infty} (\gamma \lambda)^k \delta_{t+k}
\end{equation}

\subsection{TD advantage estimator (TDAE)}

Substitute the $\lambda$-return (Eq.~\ref{lambdareturn}) into the TD update equation:
\begin{equation}\label{nTDvalueupdate}
\begin{split}
  V^{TD}(s_t) & = V_{\phi}(s_t) + \alpha (G_t^\lambda - V_{\phi}(s_t))) \\
          & = V_{\phi}(s_t) + \alpha \sum_{k=0}^{\infty} (\gamma \lambda)^{k} \delta_{s+k}
\end{split}
\end{equation}
where $\alpha$ is the update coefficient. Then, the temporal-difference advantage estimator (TDAE) can be derived:

\begin{align}\label{nTDadvantage}
     & A^{TD}(s_t, a_t) = Q^{TD}(s,a) - V^{TD}(s) \\
     & = r_{t+1} + \gamma V^{TD}(s_{t+1}) - V^{TD}(s_t) \\
     & = r_{t+1} + \gamma [V_{\phi}(s_{t+1}) + \alpha \sum_{k=0}^{\infty}(\gamma\lambda)^{k}\delta_{k+t+1}] \\
     & - [V_{\phi}(s_t) + \alpha \sum_{k=0}^{\infty}(\gamma\lambda)^{k}\delta_{k+t}] \\
     & = r_{t+1} + \gamma V_{\phi}(s_{t+1}) - V_{\phi}(s_t) \\
     & + \gamma \alpha \sum_{k=0}^{\infty}(\gamma\lambda)^{k}\delta_{t+k+1} -  \alpha \sum_{k=0}^{\infty}(\gamma\lambda)^{k}\delta_{t+k} \\
     & = \delta_t + \gamma \alpha \sum_{k=0}^{\infty}(\gamma\lambda)^{k}\delta_{t+k+1} -  \alpha \sum_{k=0}^{\infty}(\gamma\lambda)^{k}\delta_{t+k} \\
     & = (1 - \alpha) \delta_t + \alpha (\frac{1}{\lambda}-1)\sum_{k=0}^{\infty}(\gamma\lambda)^{k+1}\delta_{t+k+1}
%                  & = (1- \alpha)r_{t+1} + \alpha\gamma r_{t+2} - (1-\alpha)V_{\phi}(s_t) + \gamma(1-2\alpha)V_{\phi}(s_{t+1}) + \alpha \gamma^2 V_{\phi}(s_{t+2})
\end{align}
Comparing the form of TDAE (Eq.~\ref{nTDadvantage}) and GAE (Eq.~\ref{gae}), the essential difference is the weight distribution of the TD errors ($\delta$) in the given episode. Besides the discount weight distribution in GAE, TDAE also assigns a sliding weight of the current TD error $\delta_t$ and the subsequent TD errors ($\delta_{t+1}\rightarrow \delta_{\infty}$) based on $\alpha$. Thus the derivation of TDAE has contained the update of the value function from $V_{\phi}(s_t)$ (predicted by the value network) to $V^{TD}(s_t)$ (predicted by TD($\lambda$)). However the calculation of GAE is just based on the current $V_{\phi}(s_t)$. Therefore, TDAE contains more advanced information about the environment than GAE. The drawback of TD prediction is the model may be unstable when $\alpha$ is inappropriate. In next subsection, we will fuse GAE and TDAE to enhance the robustness and accuracy of the model.

\subsection{Dual-track advantage estimator (DTAE)}
In the next section, we will introduce entropy term to the reward function to increase the exploration opportunity of the agent in the early training stage. In this case, the agent tends to adopt the action with higher randomness rather than the greed action with maximal return predicated by the current policy. Hence the increment speed of the cumulative return curve is relatively slow in this phase and one of the effective ways to tackle this problem is to accelerate the convergence of value functions. In practical, TD methods have been found to converge faster than other update strategies such as constant-$\alpha$ Monte Carlo and dynamic programming~\cite{schulman2015high}. Nevertheless, the inappropriate selection of $\alpha$ may cause the unstable value functions or policy function in the parameters update process, but the update of GAE is more cautious. Therefore, integrating the advantages of TDAE and GAE, we propose the dual-track advantage estimator in this subsection. In the $k$th iteration, the current value network is $V_{\phi_k}$ (corresponding to the current policy) and the value network before the update is $V_{\phi_{k-1}}$ (corresponding the shadow policy). Consider the episode $\tau |{\pi_{\theta_k}} = [s_0, a_0, r_1. s_2, a_2, ...]$, which is sampled by the current policy $\pi_{\theta_k}$, we can compute the GAE and TDAE based on $V_{\phi_k}$ and $V_{\phi_{k-1}}$, respectively:

\begin{equation}\label{second}
  \begin{split}
      & \delta^{\phi_{k}}_t = r_{t+1} + \gamma V_{\phi_{k}}(s_{t+1}) - V_{\phi_{k}}(s_{t}) \\
      & A^{G}(s_t, a_t) = \sum_{k=0}^{\infty} (\gamma \lambda)^k \delta^{\phi_{k}}_{t+k} \\
      & \delta^{\phi_{k-1}}_t = r_{t+1} + \gamma V_{\phi_{k-1}}(s_{t+1}) - V_{\phi_{k-1}}(s_{t}) \\
      & A^{TD}(s_t, a_t) = (1 - \alpha) \delta^{\phi_{k-1}}_t + \alpha (\frac{1}{\lambda}-1)\sum_{k=0}^{\infty}(\gamma\lambda)^{k+1}\delta^{\phi_{k-1}}_{t+k+1}
  \end{split}
\end{equation}

And we define the dual-track advantage estimator as:

\begin{equation}\label{DTAE}
  A^{DT}(s_t, a_t) = mean \left[ A^{G}(s_t, a_t), A^{TD}(s_t,a_t) \right]
\end{equation}
There are some alternatives in combining $A^{G}$ and $A^{TD}$, but we find that the average function achieves the best results overall in the experiments (Subsection~\ref{para}).
\section{Soft policy optimization using DTAE}
In the initial training stage of our policy model, we always expect that the policy distribution is uniform so that each state can be traversed as much as possible. That is, the algorithm should be breadth-first to avoid local optimal solutions. Entropy can measure the uniformity of the policy distribution (Entropy increases as the uniformity of the policy distribution increases, and it reaches a maximum when the distribution is uniform). Hence we can define the maximum entropy objective as~\cite{ziebart2010modeling,ahmed2018understanding}:
\begin{equation}\label{entropy goal}
  J^H (\pi) = \mathbb{E}_{(s_t, a_t)\sim \rho(s_0), \pi, p}[\sum_{t=0}^{\infty} \gamma^t (r(s_t, a_t) + \eta H(\pi(s_t)))]
\end{equation}
where $H(\pi(s_t))$ is the Shannon entropy of the policy distribution $\pi(s_t)$: $ H_{\pi_{s_t}} = \int_{a} \pi(a|s) \log \pi(a|s)$ \footnote{when the action space is discrete, the entropy of policy distribution is $H_{\pi}(s_t) = \sum_{a} \pi(a|s) \log \pi(a|s)$}, $\eta$ is the temperature parameter which determines the relative importance of the entropy term against the reward~\cite{haarnoja2018soft}.

We just slight modify the definition of reward: $r^H(s_t, a_t) \doteq r(s_t, a_t) + \eta \mathbb{E}_{s_{t+1} \sim P} H(\pi(s_{t+1}))$, then the new entropy-based functions of $V^H(s_t), Q^H(s_t, a_t), A^H(s_t, a_t)$ can be established according Eq.~\ref{value function}:
\begin{equation}\label{Evalue}
\begin{split}
& V^H_{\pi}(s_t) = \mathbb{E}_{a_t, s_{t+1}, ... \sim \pi, p} [\sum_{l=0}^{\infty} \gamma^l r^H(s_{t+l})] \\
& Q^H_{\pi}(s_t, a_t) = \mathbb{E}_{s_{t+1},a_{t+1}, ... \sim p, \pi} [\sum_{l=0}^{\infty} \gamma^l r^H(s_{t+l}, a_{t+l})] \\
& A^H(s_t, a_t) = Q^H_{\pi}(s_t, a_t) - V_{\pi}^H(s_t)
\end{split}
\end{equation}
Using the entropy and entropy-based advantage function, we can quantify the performance difference between two policy as:
\begin{theorem}
Consider two policy $\hat{\pi}$ and $\pi$, let $T_{\pi}(s_t,a_t) = A^H_{\pi}(s_t, a_t) + \eta [H(\hat{\pi}(s_{t+1})) - H(\pi(s_{t+1}))]$, the superiority of $\hat{\pi}$ over $\pi$ is (See Appendix~\ref{difference} for proof):
\begin{equation}\label{superiority}
  J(\hat{\pi}) - J(\pi) = \mathbb{E}_{\tau|\hat{\pi}} \gamma^t T_{\pi} (s_t,a_t)
\end{equation}
\end{theorem}

Let $\hat{T}(s_t) = \mathbb{E}_{a\sim \hat{\pi}(\cdot|s_t)} T_{\pi} (s_t,a_t)$, Eq.~\ref{superiority} can be transformed into:
\begin{equation}\label{jreturn}
J(\hat{\pi}) = J(\pi) + \mathbb{E}_{\tau|\hat{\pi}} \gamma^t \hat{T}(s_t)
\end{equation}

In practice, it is difficult to optimize Eq.~\ref{jreturn} directly since the heavy dependency of $\tau$ on $\hat{\pi}$, thus we can replace $\tau|\hat{\pi}$ into $\tau|\pi$ to approximate Eq.~\ref{jreturn} to simplify the optimization process:
\begin{equation}\label{sreturn}
L_{\pi}(\hat{\pi}) = J(\pi) + \mathbb{E}_{\tau|\pi} \gamma^t \hat{T}(s_t)
\end{equation}
\begin{equation}\label{s}
r^H(s_t, a_t) \doteq r(s_t, a_t) + \eta \mathbb{E}_{s_{t+1} \sim P} H(\pi(s_{t+1}))
\end{equation}
The difference of Eq.~\ref{jreturn} and Eq.~\ref{sreturn} is that $s_t$ is sampled by $\pi$ or $\hat{\pi}$. Note that $J(\hat{\pi})$ and $L_{\pi}(\hat{\pi})$ are both differentiable functions about the parameter vector $\theta$, and we have:
\begin{equation}\label{cond}
  \begin{split}
      & J(\pi_{\theta_{old}}) = L_{\pi_{\theta_{old}}}(\pi_{\theta_{old}}) \\
      & \nabla_{\theta} J(\pi_{\theta}) |_{\theta = \theta_{old}} = \nabla_{\theta} L_{\pi_{\theta_{old}}}(\pi_{\theta}) |_{\theta = \theta_{old}}
  \end{split}
\end{equation}
where $\theta_{old}$ denotes the parameter vector of the current policy and $\hat{\theta}$ denotes the parameter vector of the new (expected) policy. That implies the sufficient small step $\theta_{old} \rightarrow \theta_{old} + \Delta{\theta}$ which we take to improve $L$ at $\theta_{old}$ will also improve $J$. In practice, however, the optimal policy parameter vector solved by Eq.~\ref{sreturn} is $\hat{\theta} = \arg \max_{\hat{\theta}}L_{\theta_{old}}(\hat{\theta})$ and the step $|\hat{\theta} - \theta_{old}| \gg\Delta \theta$. In order to satisfy the restriction of Eq.~\ref{cond} (the step is sufficient small), \cite{schulman2015trust} proposed the coupled policy to solve this problem:

\begin{definition}
Define the indicator variable $\mathbb{I}_s = 1$ if $a = \hat{a} |s$, else $\mathbb{I}_s = 0$, where $a$ is sampled by $\pi$ and $\hat{a}$ is sampled by $\hat{\pi}$.  $(\pi, \hat{\pi})$ is a $\kappa$-coupled policy if $p(\mathbb{I}_s = 0) \leq \kappa$ for all $s$.
\end{definition}
Consider the coupled policy ($\pi, \hat{\pi}$), we can derive the low bound of $J(\hat{\pi})$ as follows\footnote{For convenience: $\pi = \pi(\theta), \hat{\pi} = \pi(\hat{\theta}), \pi_{old} = \pi(\theta_{old})$}(See Appendix~\ref{pbound} for proof):
\begin{equation}\label{bound}
  J(\hat{\pi}) \geq L_{\pi_{old}}(\hat{\pi}) - C D_{KL}^{\max}(\pi_{old}, \hat{\pi})
\end{equation}
where $C = \frac{2 \xi \gamma}{(1-\gamma)^2}$, $\xi = 2 \max_{s,a}|A^H_{\pi}(s,a)| + \frac{\eta}{2\kappa}\log e + \frac{\delta\eta}{\kappa}$, $D_{KL}^{\max} = \max_s D_{KL}(\pi_{old}(\cdot|s) || \hat{\pi}(\cdot|s))$ and $D_{KL}$ is the KL diverge. Eq.~\ref{bound} shows that we can guarantee $J$ is non-decreasing if we maximize the right part at each iteration. That means we can improve the performance of policy by solve the following optimization issue:
\begin{equation}\label{maxobject}
  \max_{\hat{\pi}} \left[ L_{\pi_{old}}(\hat{\pi}) - C D_{KL}^{\max}(\pi_{old}, \hat{\pi}) \right]
\end{equation}
Finally, Eq.~\ref{maxobject} can be simplified as follows (See Appendix~\ref{converge} for proof):
\begin{equation}\label{trobejct}
  \begin{split}
      & \hat{\mathbb{E}}_t \left[ \frac{\pi_{\hat{\theta}}(a_t|s_t)}{\pi_{old}(a_t|s_t)} T_{\pi_{old}}(s_t, a_t) \right] \\
      & s.t. \quad \hat{\mathbb{E}}_t [D_{KL}(\pi_{old}(\cdot|s) || \hat{\pi}(\cdot|s))] \leq \delta
  \end{split}
\end{equation}
We have discussed the drawbacks of the trust region method above, and thus we apply the clip trick to Eq.~\ref{trobejct}. Define probability ratio $r_t = \pi_{\hat{\theta}}(a_t|s_t) / \pi_{\theta_{old}}(a_t|s_t)$, we have:
\begin{equation}\label{cobjective}
 \max_{\hat{\theta}} \quad \mathbb{\hat{E}}_t \left [\min \left( r_t T_t, clip(r_t, 1-\epsilon, 1+\epsilon) T_t \right) \right]
\end{equation}
where $T_t = A_{\pi_{old}}^{H}(s_t,a_t) + \eta [H(\hat{\pi}(s_{t+1})) - H(\pi_{old}(s_{t+1})) ]$. In experiment, we replace $A_{\pi_{old}}^{H}$ by DTAE to enhance the robustness and accuracy of the algorithm. Consider the current entropy-based value network $V_{\pi_{old}}^H$ and $V_{\pi_{sold}}^H$ ($\pi_{sold}$ is the shadow policy of $\pi_{old}$). The TD errors are:
 \begin{equation}\label{sTD}
   \begin{split}
       & \delta^{\pi_{old}}_t = r^H_{t+1} + \gamma V^H_{\pi_{old}}(s_{t+1}) - V^H_{\pi_{old}}(s_{t}) \\
       & \delta^{\pi_{sold}}_t = r^H_{t+1} + \gamma V^H_{\pi_{sold}}(s_{t+1}) - V^H_{\pi_{sold}}(s_{t})
   \end{split}
 \end{equation}
Therefore, according to Eq.~\ref{second}, $A^{G}(s_t, a_t)$ and $A^{TD}(s_t, a_t)$ can be calculated by $\delta^{\pi_{old}}_t$ and $\delta^{\pi_{sold}}_t$ respectively, and the final advantage function is:
\begin{equation}\label{T}
\begin{split}
    & A^H_t = mean \left[A^{G}(s_t, a_t), A^{TD}(s_t, a_t) \right] \\
    & T_t = A^H_t + \eta [H(\hat{\pi}(s_t)) - H(\pi_{old}(s_t))
\end{split}
\end{equation}
%\begin{figure*}
%  \centering
%  \includegraphics[width=14cm]{Comp_Methods.eps}
%  \caption{Training curves on the Mujoco continuous control tasks. These curves reflect the change of cumulative return over 1 million time steps. The solid line denotes the average of 10 trails generated by random time seeds and the shaded region is bounded by the maximum and minimum of the 10 trails.}\label{comp_methods}
%\end{figure*}
\begin{figure*}
  \begin{subfigure}{.33\textwidth}
  \centering
  \includegraphics[width=.8\linewidth]{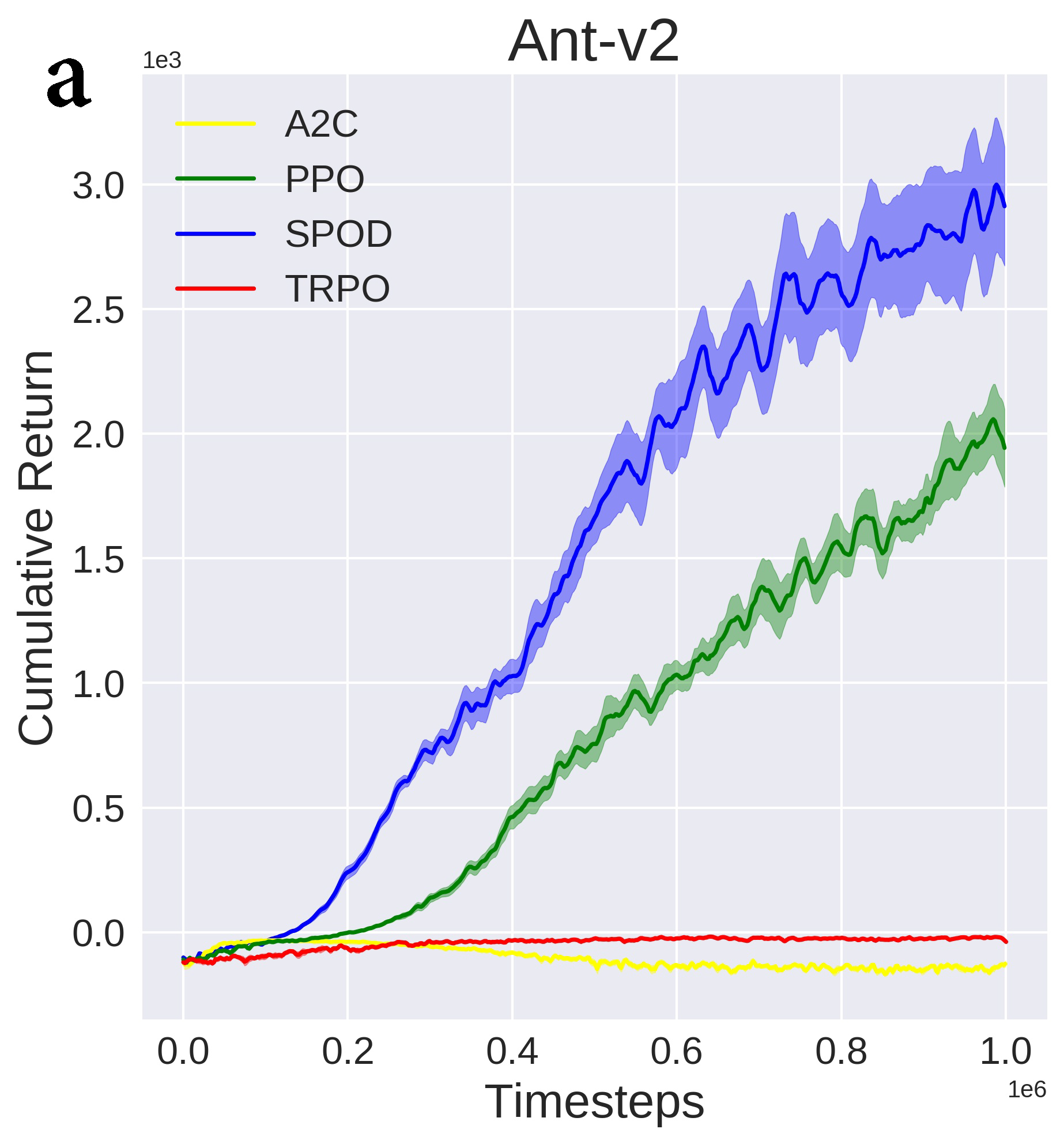}
  \end{subfigure}%
  \begin{subfigure}{.33\textwidth}
  \centering
  \includegraphics[width=.8\linewidth]{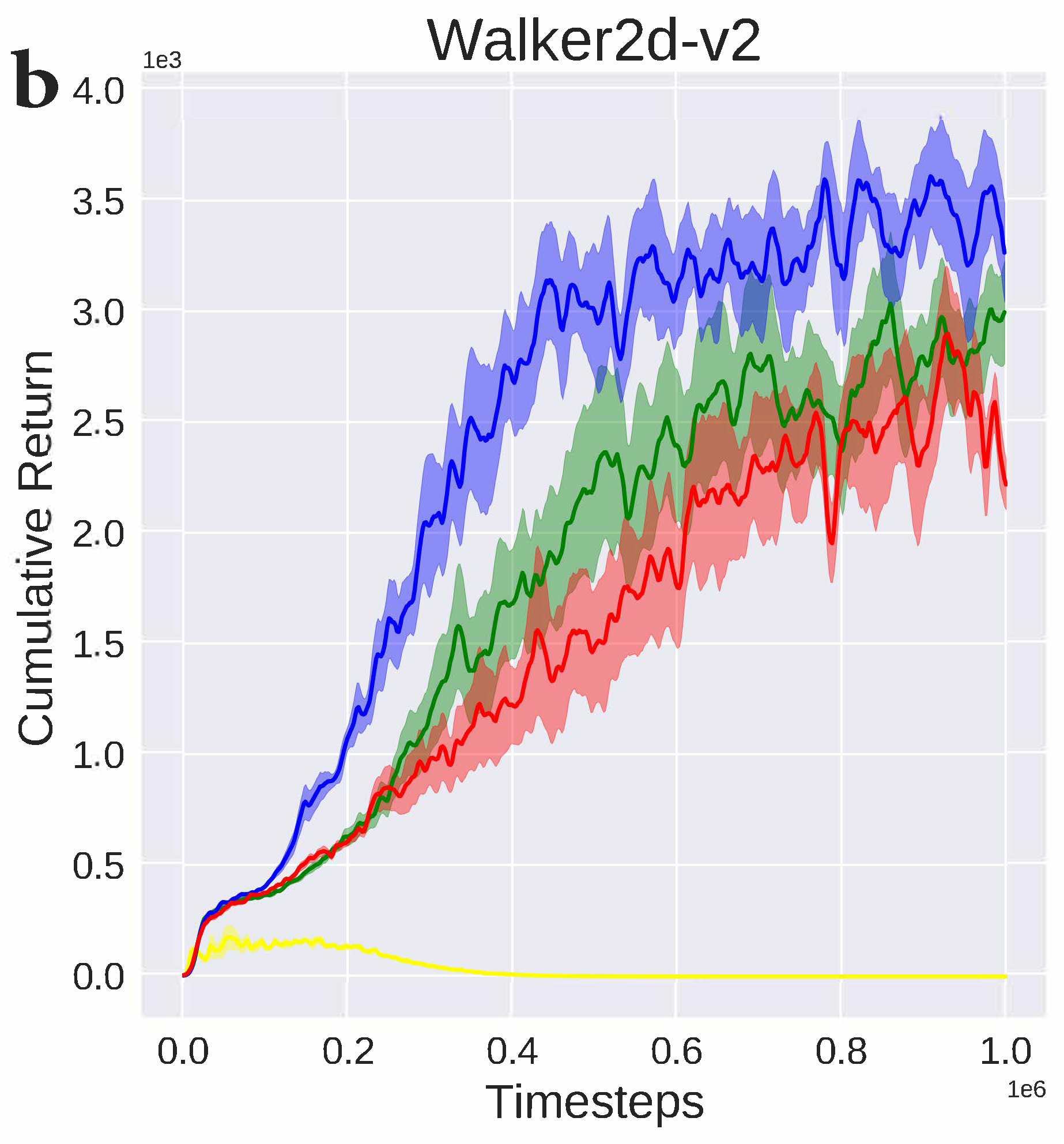}
  \end{subfigure}
  \begin{subfigure}{.33\textwidth}
  \centering
  \includegraphics[width=.8\linewidth]{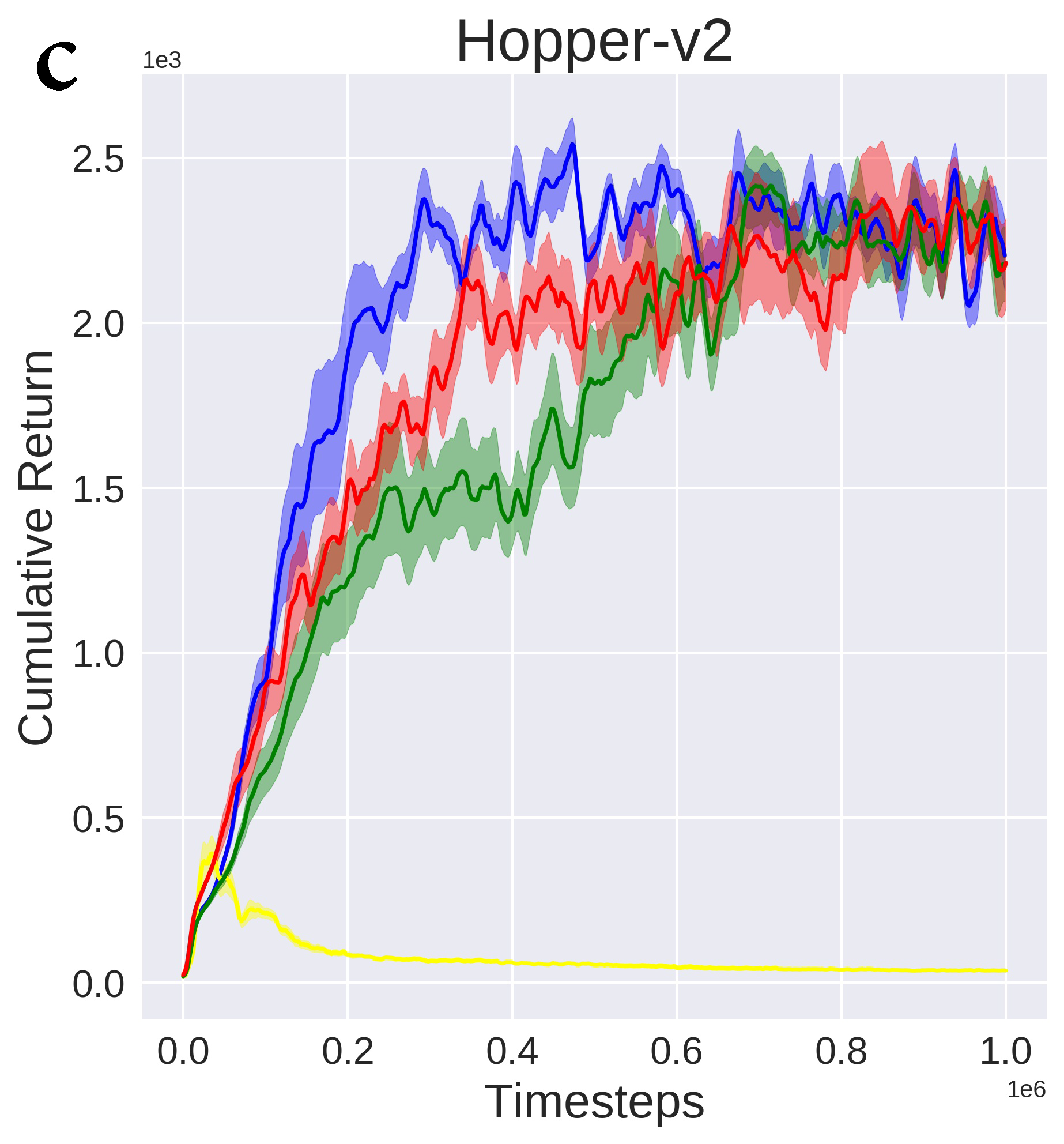}
  \end{subfigure} \\
  \begin{subfigure}{.33\textwidth}
  \centering
  \includegraphics[width=.8\linewidth]{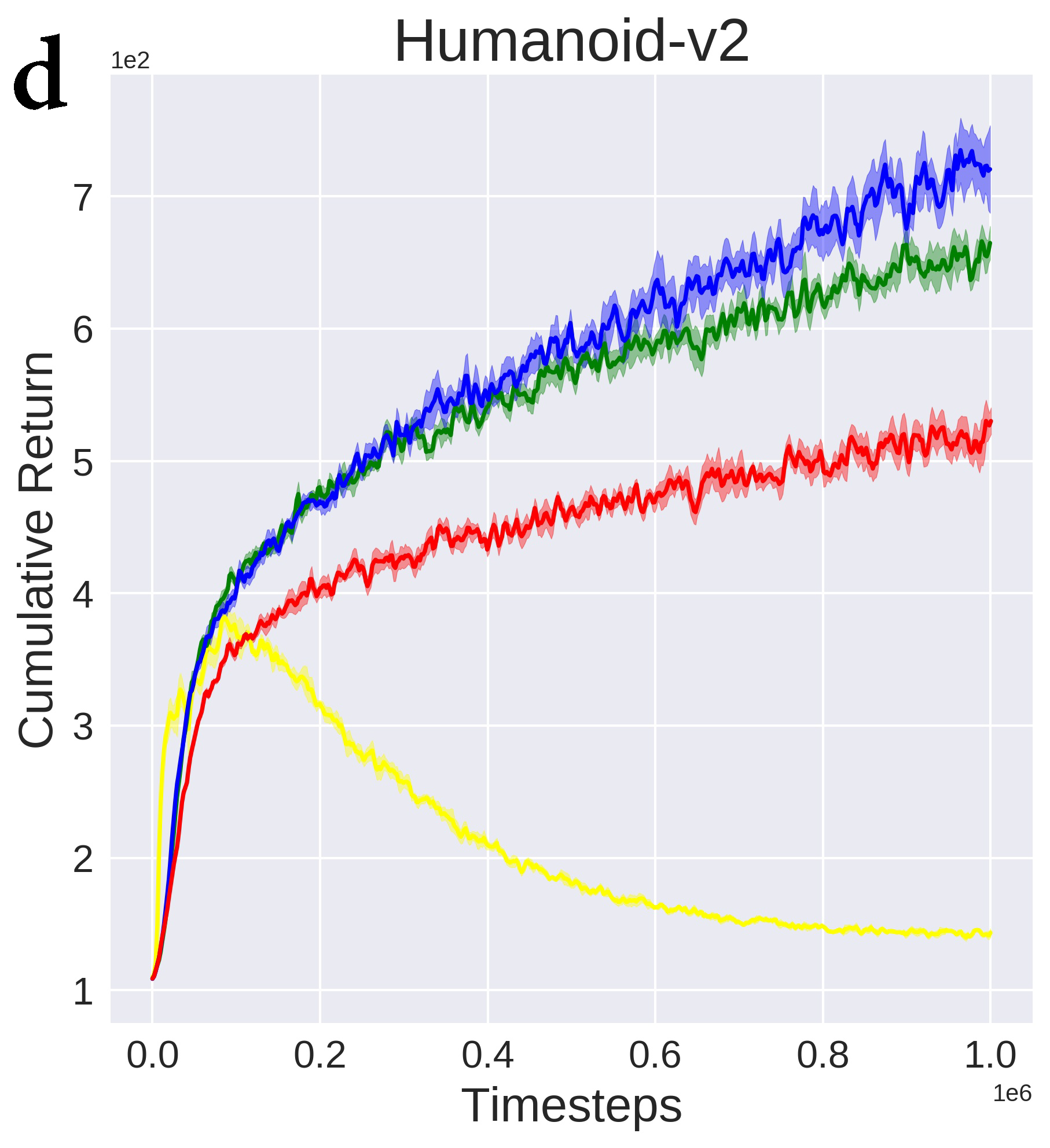}
  \end{subfigure}%
  \begin{subfigure}{.33\textwidth}
  \centering
  \includegraphics[width=.8\linewidth]{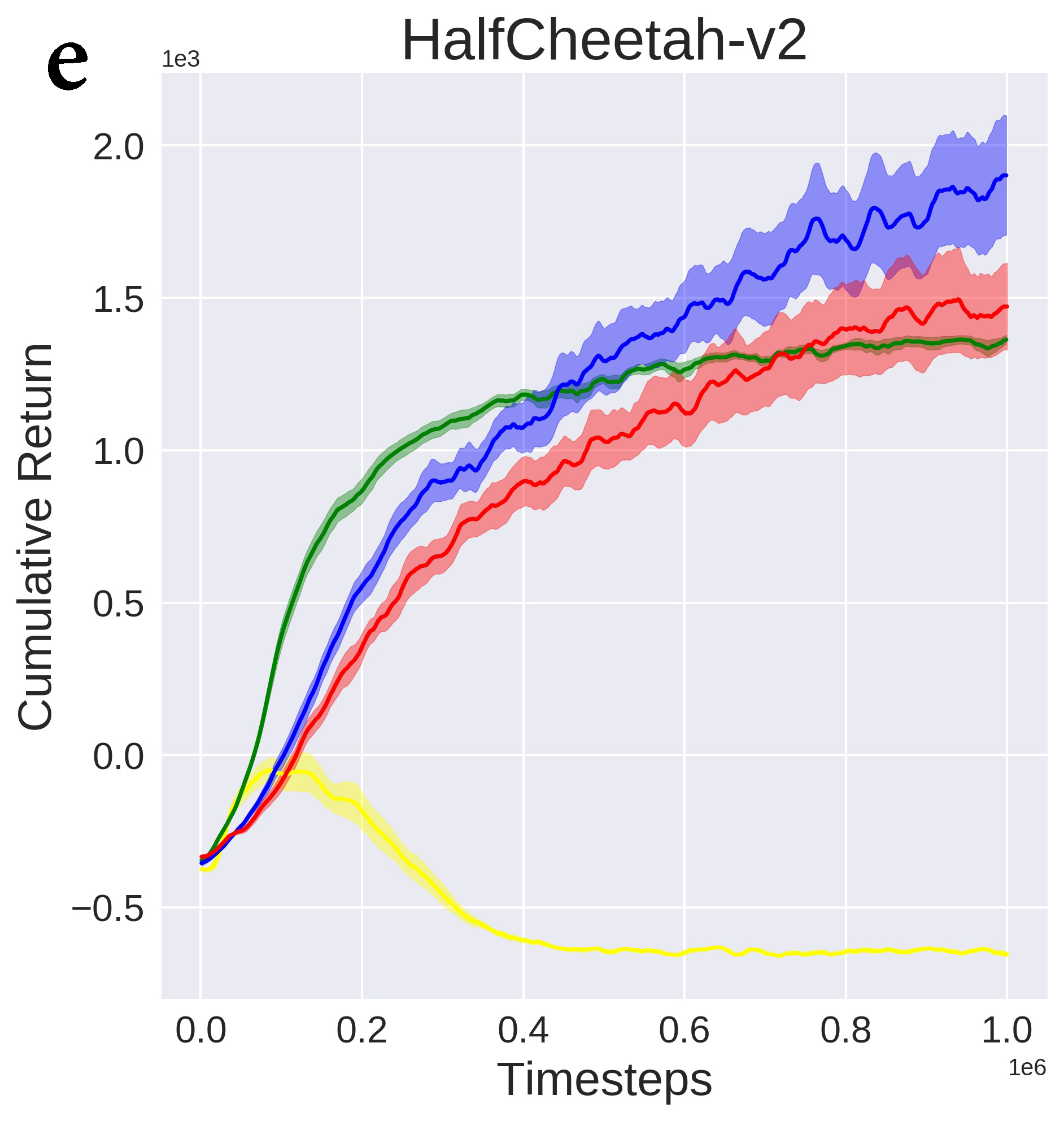}
  \end{subfigure}
  \begin{subfigure}{.33\textwidth}
  \centering
  \includegraphics[width=.8\linewidth]{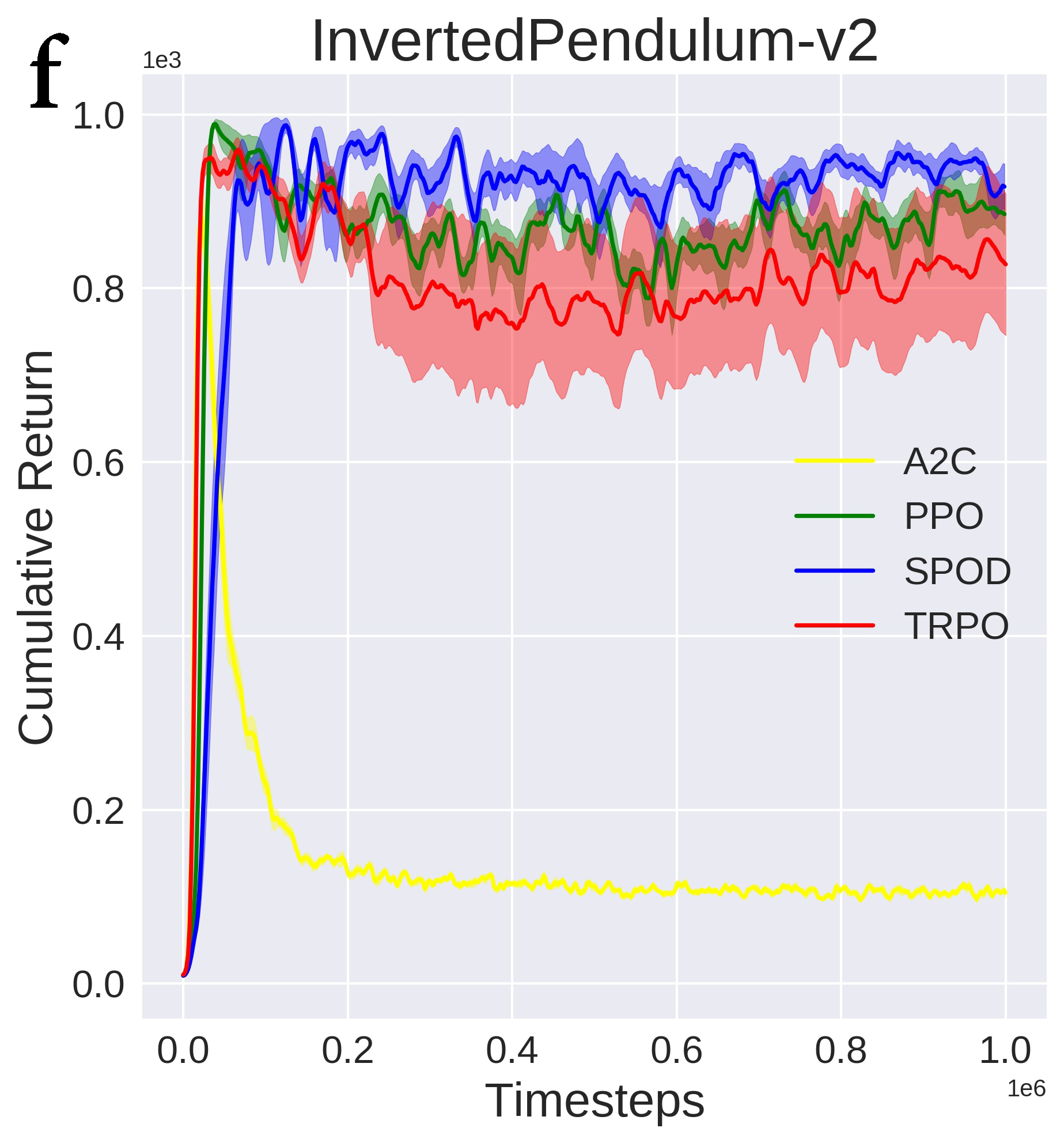}
  \end{subfigure}
  \caption{Training curves on the Mujoco continuous control tasks. These curves reflect the change of cumulative return over 1 million time steps. The solid line denotes the average of 10 trails generated by random time seeds and the shaded region is bounded by the maximum and minimum of the 10 trails.}\label{comp_methods}
\end{figure*}
%\begin{algorithm}
%\caption{SPOD}\label{SPOD}
%\textbf{Input:} initialize policy parameters $\theta_0$, shadow policy parameters $\tilde{\theta}_0$, value function parameters $\phi_0$ and shadow value function $\tilde{\phi}_0$.
% \begin{algorithmic}[1]
%   \FOR {$k = 0$ to $K$}
%   \STATE Collect set of trajectories $D_{k} = \{\tau_{i}\}$ by running policy $\pi_{k} = \pi(\theta_{k})$ in the environment.
%   \STATE Compute rewards-to-go $G_t$ based on $D_k$
%   \STATE Compute the TD errors, $\delta^{\theta_k}_t$ and $\delta^{\tilde{\theta}_k}_t$ based on $V_{\phi_k}$, $V_{\tilde{\phi}_{k}}$ and $\pi_{\theta_k}$, $\pi_{\tilde{\theta}_k}$ respectively (Eq.~\ref{sTD}). And then compute the entropy-based advantage function $T_t$ according to Eq.~\ref{T}.
%   \STATE $\tilde{\theta}_{k+1}  = \theta_k$, $\tilde{\phi}_{k+1} = \phi_k$
%   \STATE Update the policy by maximizing the PPO-Clip objective:
%          \begin{equation*}
%          \begin{split}
%              & \theta_{k+1} = \arg \max_{\theta} \frac{1}{|D_{k}|} \sum_{\tau \in D_k} \sum_{t = 0} \\
%              & \min(\frac{\pi_\theta(a_t|s_t)}{\pi_{\theta_k}(a_t|s_t)} T_t, g(\epsilon, T_t))
%          \end{split}
%          \end{equation*}
%   \STATE Fit value function by regression on mean-squared error:
%          \begin{equation*}
%            \phi_{k+1} = \arg \min_{\phi} \frac{1}{|D_{k}|T} \sum_{\tau \in D_k} \sum_{t = 0}^{T}(G^H_t - V^H_{\phi_{k}}(s_t))^2
%          \end{equation*}
%   \ENDFOR
% \end{algorithmic}
%\end{algorithm}
\section{Experiments and results}
To evaluate the performance of RL algorithms, in this paper, we use the OpenAI gym benchmark suite, which engineered by Mujoco, to simulate the continuous robotic control environments. In particular, the high-dimensional control tasks such as Ant-v2, Walker-v2, Humanoid-v2 are challenging for agents. There are some criteria to judge the performance of agents in one task: cumulative return (the mean of the training curves), training speed (the growth rate of the training curves) and stability (the shaded region of the training curves). High return shows the tested algorithm is effective, fast training speed means the corresponding algorithm has the efficient sample utilization capacity, and small shaded region indicates the corresponding agent can achieve similar results under fluctuating initial conditions. Based on the above criteria, we will compare the proposed method (SPOD) with the classical on-policy algorithms using the same hyper-parameters. Meanwhile, we will test the sensibility of the proposed model to the hyper-parameters and the contribution of particular components of SPOD to the final performance. In each experiment, the figure is plotted by the mean and standard deviation of 10 trials generated by random time seeds. The default hyper-parameters are show in subsection~\ref{supply}.

\subsection{Comparative Evaluation}
In this experiment, we compare our algorithm (the corresponding code is released on GitHub\footnote{https://github.com/Code-Papers/SPOD}) against the advanced on-policy optimization methods: A2C, TRPO and PPO, as implemented by OpenAI's baselines repository\footnote{https://github.com/openai/baselines}. Fig.~\ref{comp_methods} shows, overall, the proposed algorithm is superior to other three baseline methods both in learning speed and cumulative return. That implies the introduction of dynamical entropy granted contributes to seek the more returnable trajectories since it well balances the exploration and exploitation opportunity in training. Nevertheless, the increase of randomness will cause the low training speed in the early stage. To tackle this drawback, we proposed the dual-track advantage estimator, which uses both the current value function $\phi_{k}$ and the corresponding shadow value function $\phi_{k-1}$, to accelerate the convergence of advantage function. The results of high-dimensional control tasks (Ant-v2, Walker-v2) shows DTAE has achieved our expectation and indirectly enhanced the example efficiency. In the low-dimensional control tasks (Hopper-v2, InvertedPendulum-v2), TRPO, PPO, SPOD all show excellent performance in the final stage. However, the variances (shaded region of the curve) of SPOD is significantly narrower than PPO (Fig.~\ref{comp_methods}c) and TRPO (Fig.~\ref{comp_methods}f).

\subsection{Parameters analysis}\label{para}

In this subsection, we show the effects of variable hyper-parameters on performing of SPOD through comparative experiments. Further, the tricks to determine the scale of hyper-parameters also be discussed.
%\begin{figure*}
%  \centering
%  \includegraphics[width=14cm]{Comp_Alpha.eps}
%  \caption{Comparison of different TD update coefficient $\alpha$ on the performance of SPOD. $\alpha = 0.1$ indicates the update rate of value function is slow in TD method and the agent is conservative in adopting the new explored policy, $\alpha = 0.4$ indicates the agent will adopt the compromise between new and old policies, and $\alpha =0.9$ means the agent tends to completely adopt the new explored policy.}\label{comp_alpha}
%\end{figure*}
\begin{figure*}
  \begin{subfigure}{.33\textwidth}
  \centering
  \includegraphics[width=.8\linewidth]{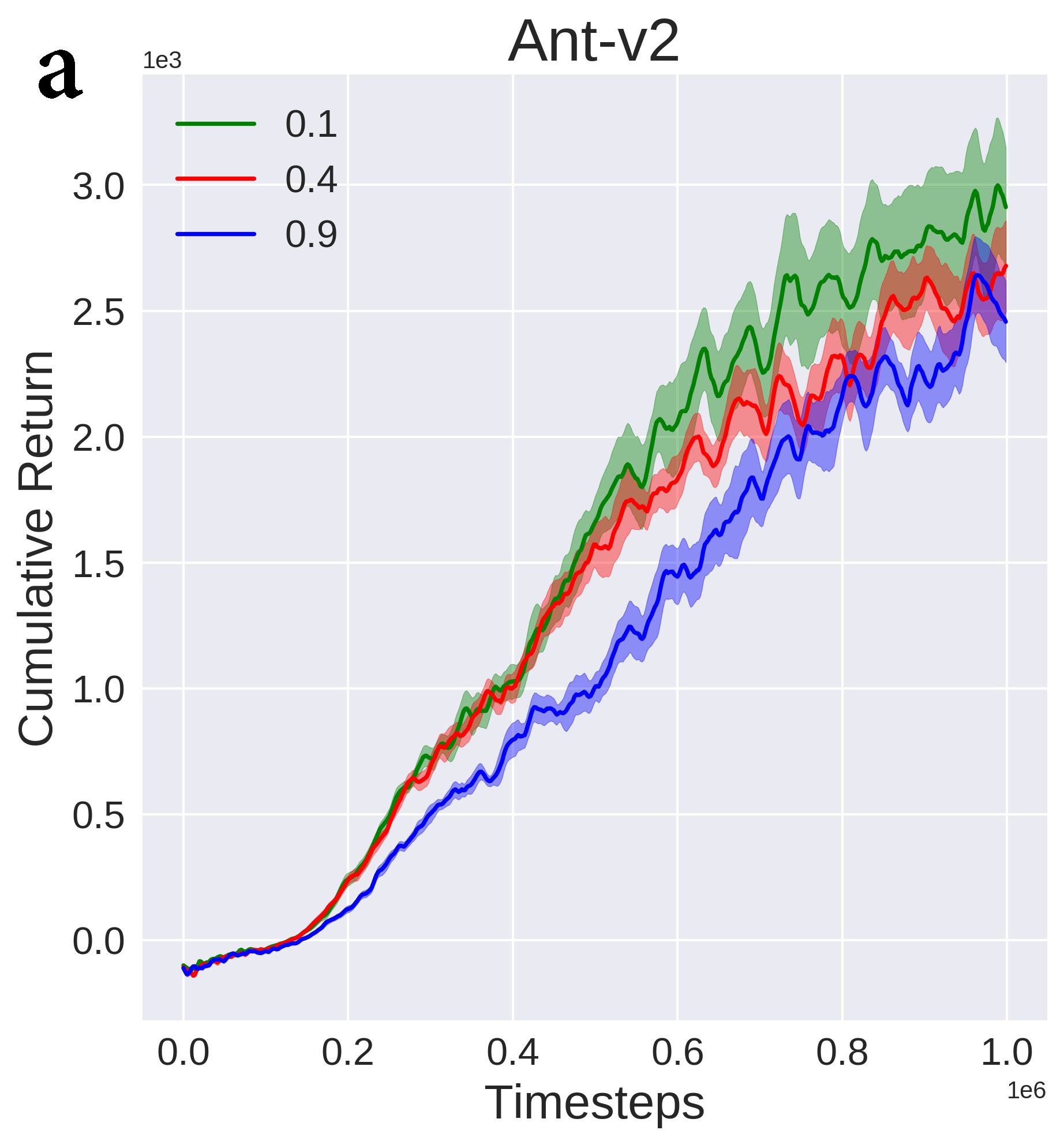}
  \end{subfigure}%
  \begin{subfigure}{.33\textwidth}
  \centering
  \includegraphics[width=.8\linewidth]{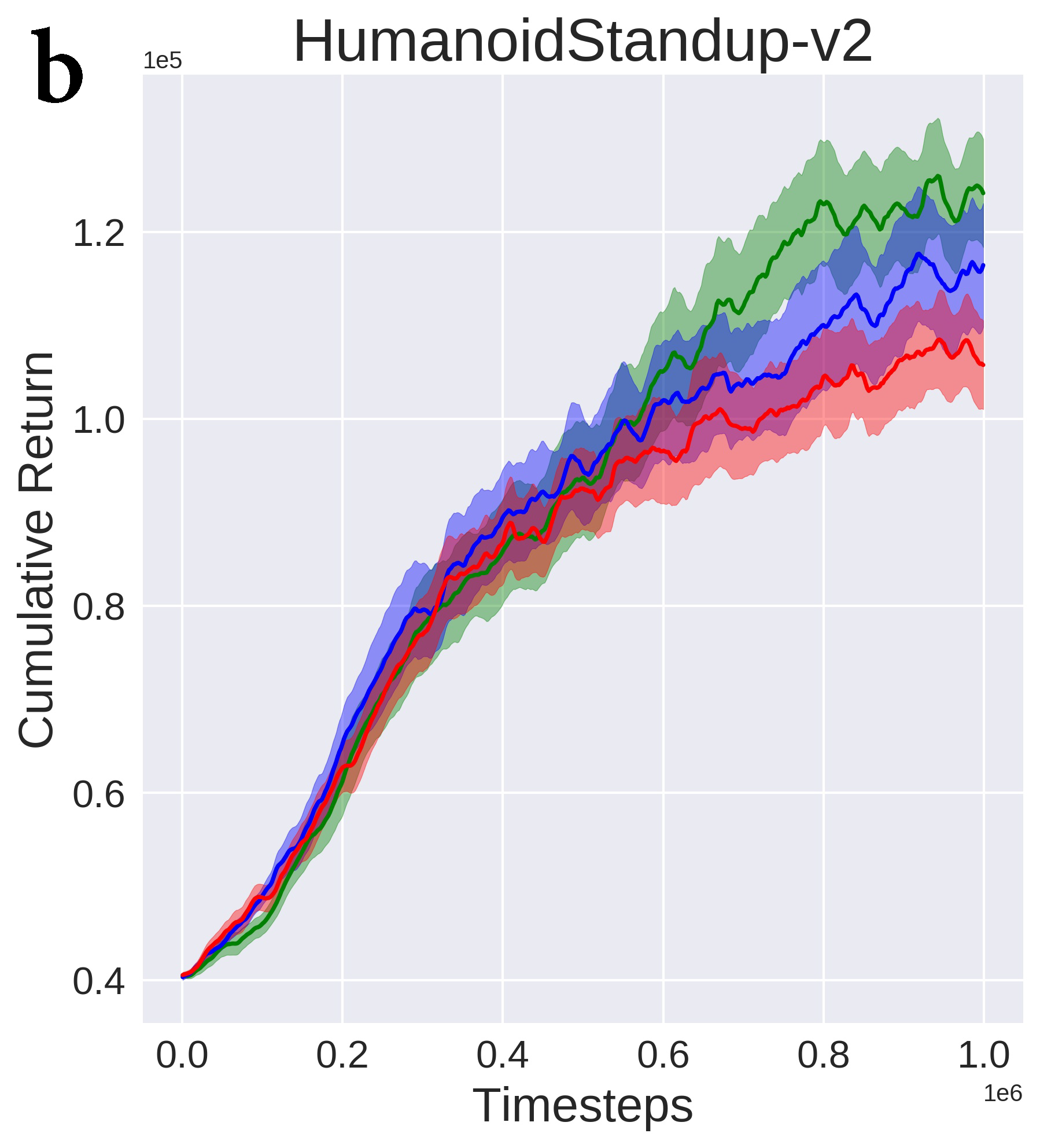}
  \end{subfigure}
  \begin{subfigure}{.33\textwidth}
  \centering
  \includegraphics[width=.8\linewidth]{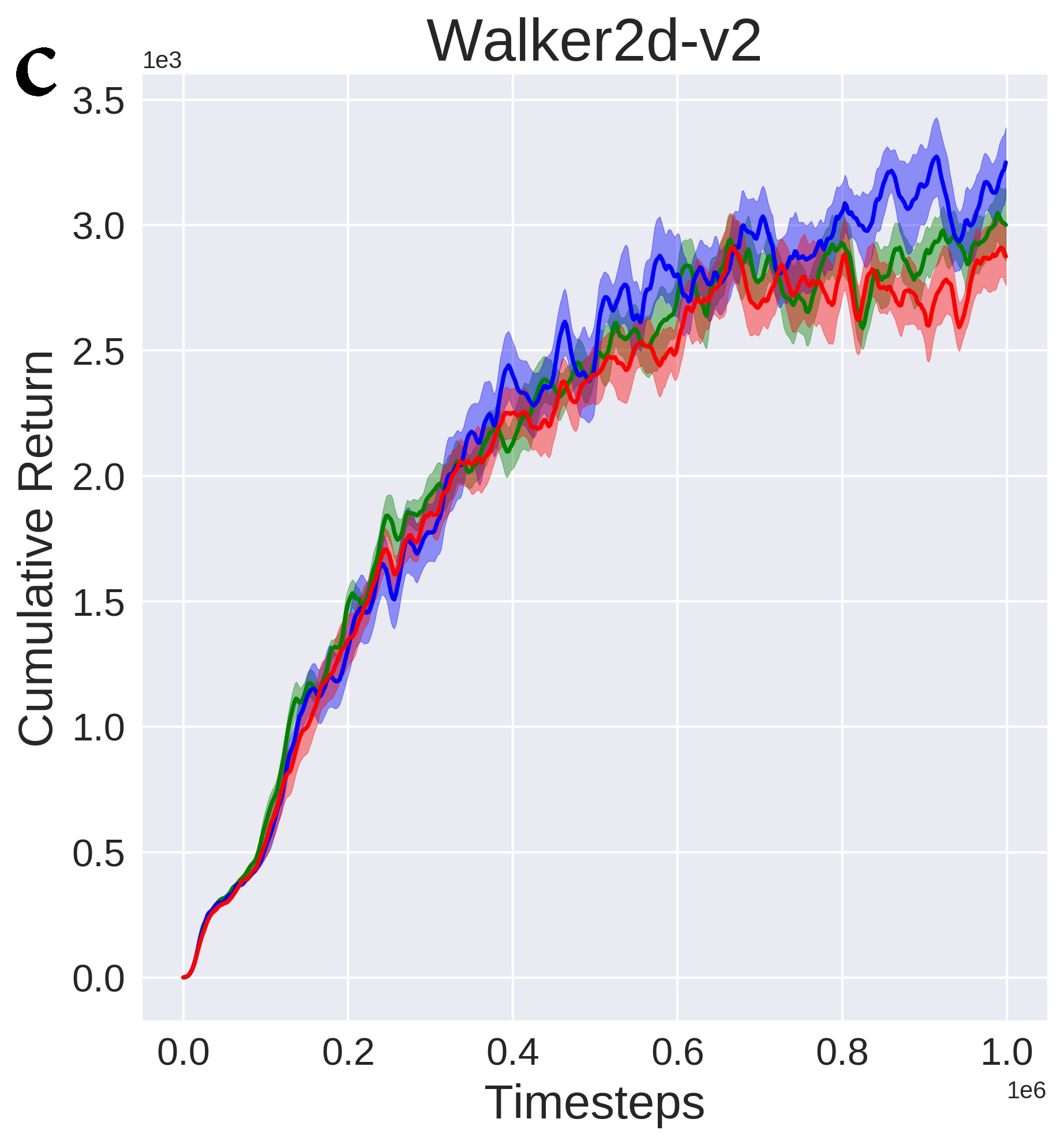}
  \end{subfigure}
  \caption{Comparison of different TD update coefficient $\alpha$ on the performance of SPOD. $\alpha = 0.1$ indicates the update rate of value function is slow in TD method and the agent is conservative in adopting the new explored policy, $\alpha = 0.4$ indicates the agent will adopt the compromise between new and old policies, and $\alpha =0.9$ means the agent tends to completely adopt the new explored policy.}\label{comp_alpha}
\end{figure*}

\textbf{TD update coefficient:} $\alpha$ determines the update ratio of value function in Bellman equation. Larger $\alpha$ indicates the agent is more inclined to adopt the new explored policy rather than the old policy. In this paper, the exploration degree is controlled by entropy coefficient and the update of shadow policy is used for increasing accuracy of advantage function. Therefore, small $\alpha$ is favorable to optimize the value functions in SPOD and the results (Fig.~\ref{comp_alpha}) are consistent with this hypothesis. Furthermore, the performances of agent keep stable under fluctuant $\alpha$, illustrating that the algorithm is tolerant in selecting the hyper-parameters. Meanwhile, the adjacent training curves under random time seeds also can reflect the stability of SPOD.

%\begin{figure*}
%  \centering
%  \includegraphics[width=14cm]{Comp_Comb.eps}
%  \caption{Comparison of different combine methods of GAE and TDAE in SPOD (Eq.~\ref{alter}). $mean$, $\max$, $\min$ denote the mean, maximum and minimum of GAE and TDAE respectively. $beta = 0.99$ denotes the weight $\beta = 0.99$ in Eq.~\ref{alter}.}\label{comp_comb}
%\end{figure*}
\begin{figure*}
  \begin{subfigure}{.33\textwidth}
  \centering
  \includegraphics[width=.8\linewidth]{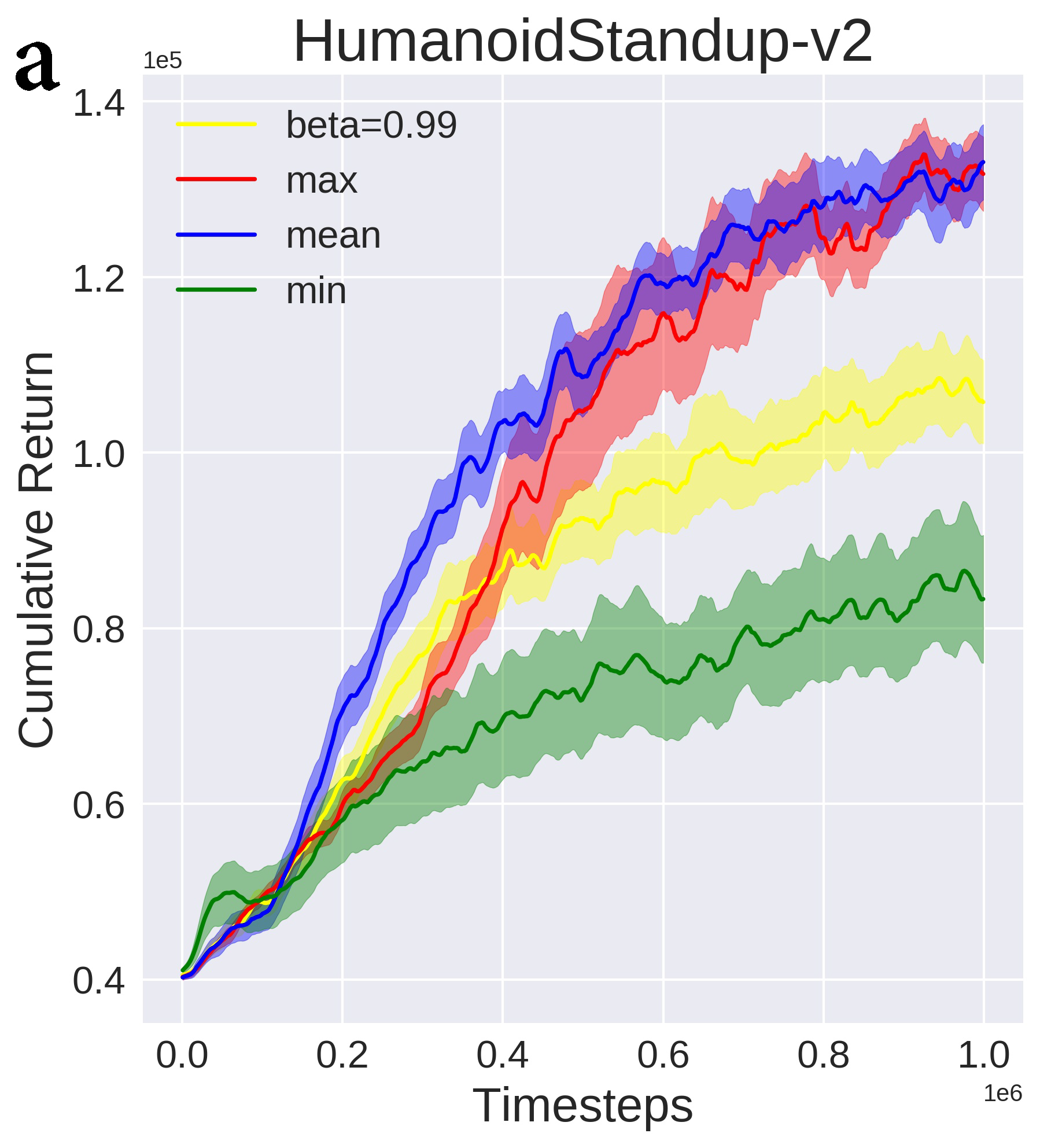}
  \end{subfigure}%
  \begin{subfigure}{.33\textwidth}
  \centering
  \includegraphics[width=.8\linewidth]{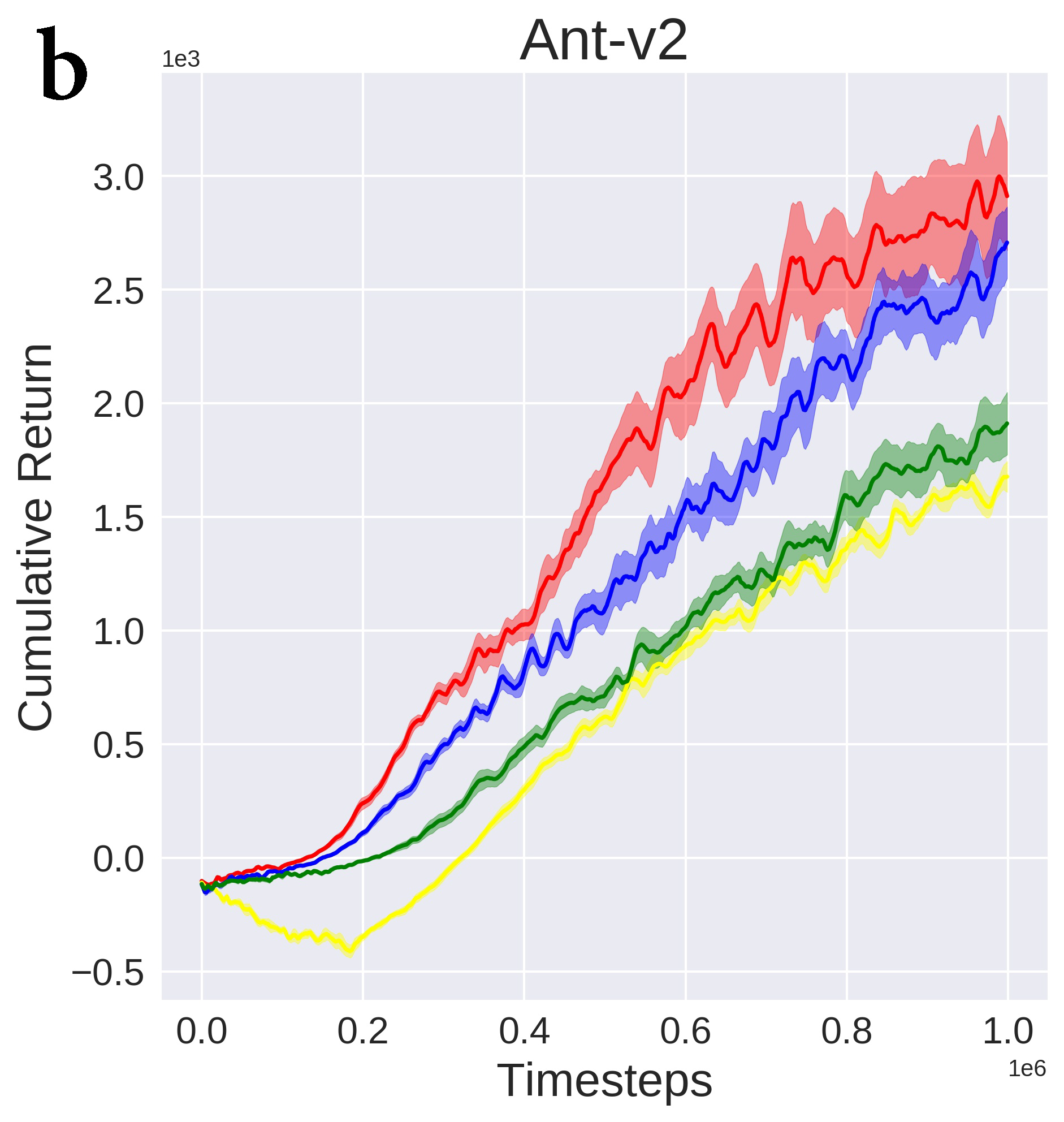}
  \end{subfigure}
  \begin{subfigure}{.33\textwidth}
  \centering
  \includegraphics[width=.8\linewidth]{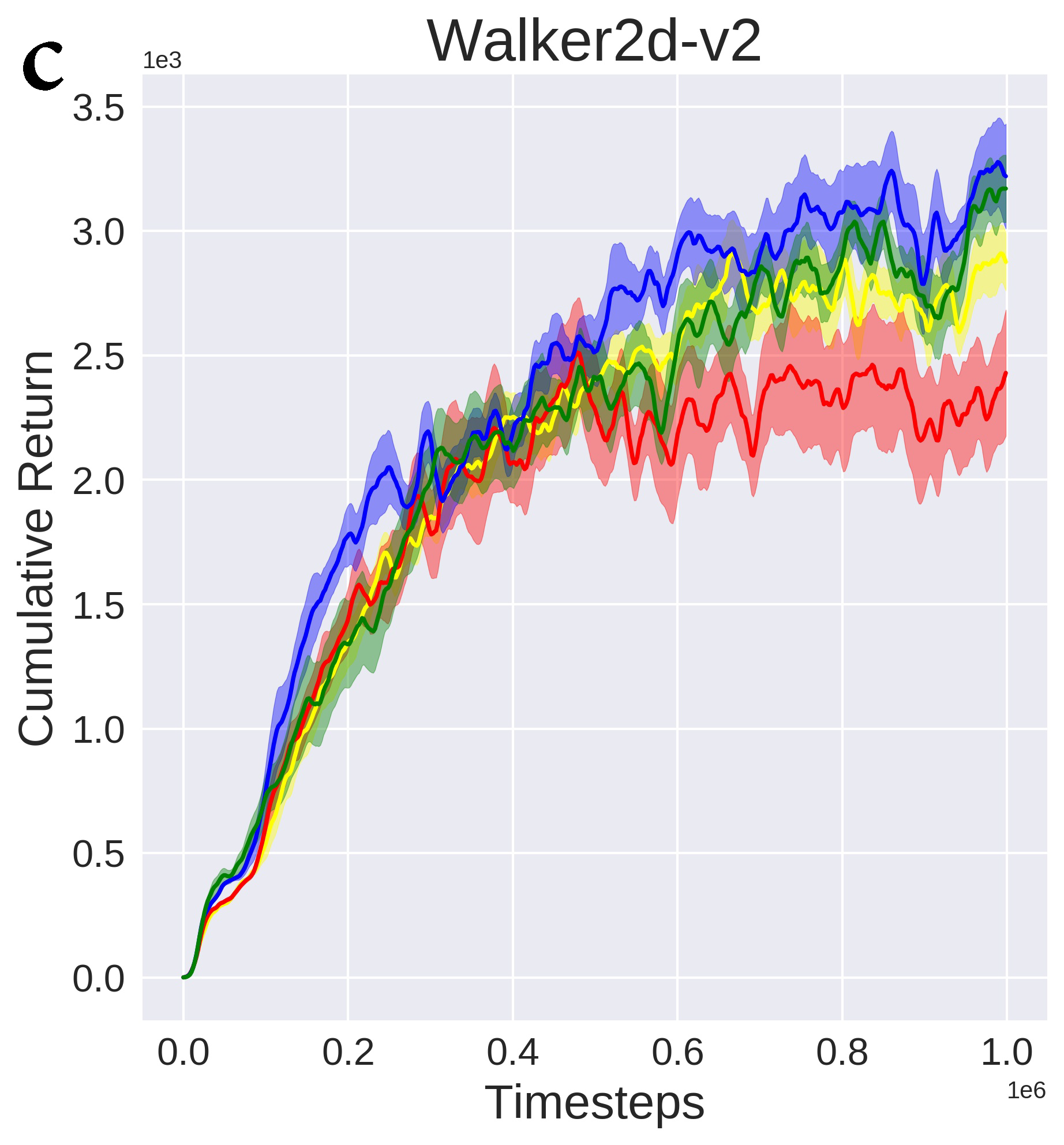}
  \end{subfigure}
  \caption{Comparison of different combine methods of GAE and TDAE in SPOD (Eq.~\ref{alter}). $mean$, $\max$, $\min$ denote the mean, maximum and minimum of GAE and TDAE respectively. $beta = 0.99$ denotes the weight $\beta = 0.99$ in Eq.~\ref{alter}.}\label{comp_comb}
\end{figure*}

\textbf{Combine methods of DTAE:} Beside the combine method of DTAE in Eq.~\ref{second}, there are some alternatives:
    \begin{equation}\label{alter}
      \begin{split}
         A^{DT} & = \max [A^G, A^{TD}] \\
             or & = \min [A^G, A^{TD}] \\
             or & = \beta A^G + (1-\beta) A^{TD}
      \end{split}
    \end{equation}
    From Fig.~\ref{comp_comb}, $mean$ performs both excellent and reliable in the three control tasks. $\max$ is also effective in Humanoid Standup-v2 and Ant-v2 but the variances of the corresponding learning curves are relative larger than $mean$, indicating $\max$ is sensitive to the fluctuant environments. Additionally, this method may cause overestimation problem in training process~\cite{fujimoto2018addressing}. Fig.~\ref{comp_comb}(a-b) show the $\min$ term suppresses the performance of SPOD from the early training stage and the results are unstable in HumanoidStandup-v2. When $\beta = 0.99$, the corresponding curves can be approximately regarded as calculated by GAE. Therefore, from Fig.~\ref{comp_comb}(a-b), we can safely conclude that DTAE not only gains high cumulative return but also has the smaller variance than GAE. The learning speed of DTAE is also significant faster than GAE, especially in the early training stage and thus it indirectly increases the sample utilization efficiency.
%\begin{figure*}
%  \centering
%  \includegraphics[width=14cm]{Comp_Ent.eps}
%  \caption{Comparison of different scales of temperature parameter $\eta$ on the performance of SPOD in three high-dimensional control tasks. The greater $\eta$ indicates the agent is likely to explore the new more returnable policies in the early training stage, and vice versa.}\label{comp_ent}
%\end{figure*}
\begin{figure*}
  \begin{subfigure}{.33\textwidth}
  \centering
  \includegraphics[width=.8\linewidth]{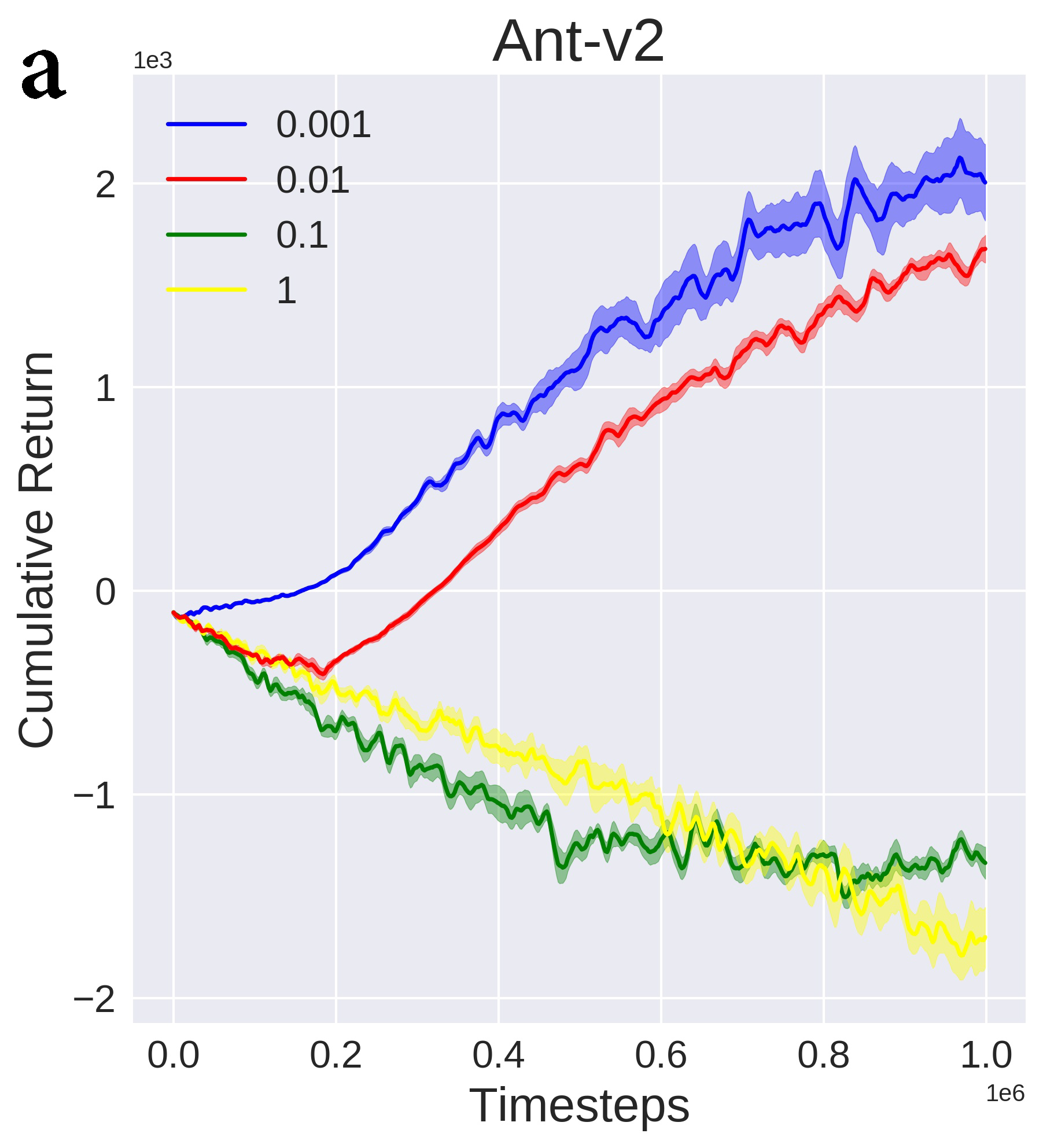}
  \end{subfigure}%
  \begin{subfigure}{.33\textwidth}
  \centering
  \includegraphics[width=.8\linewidth]{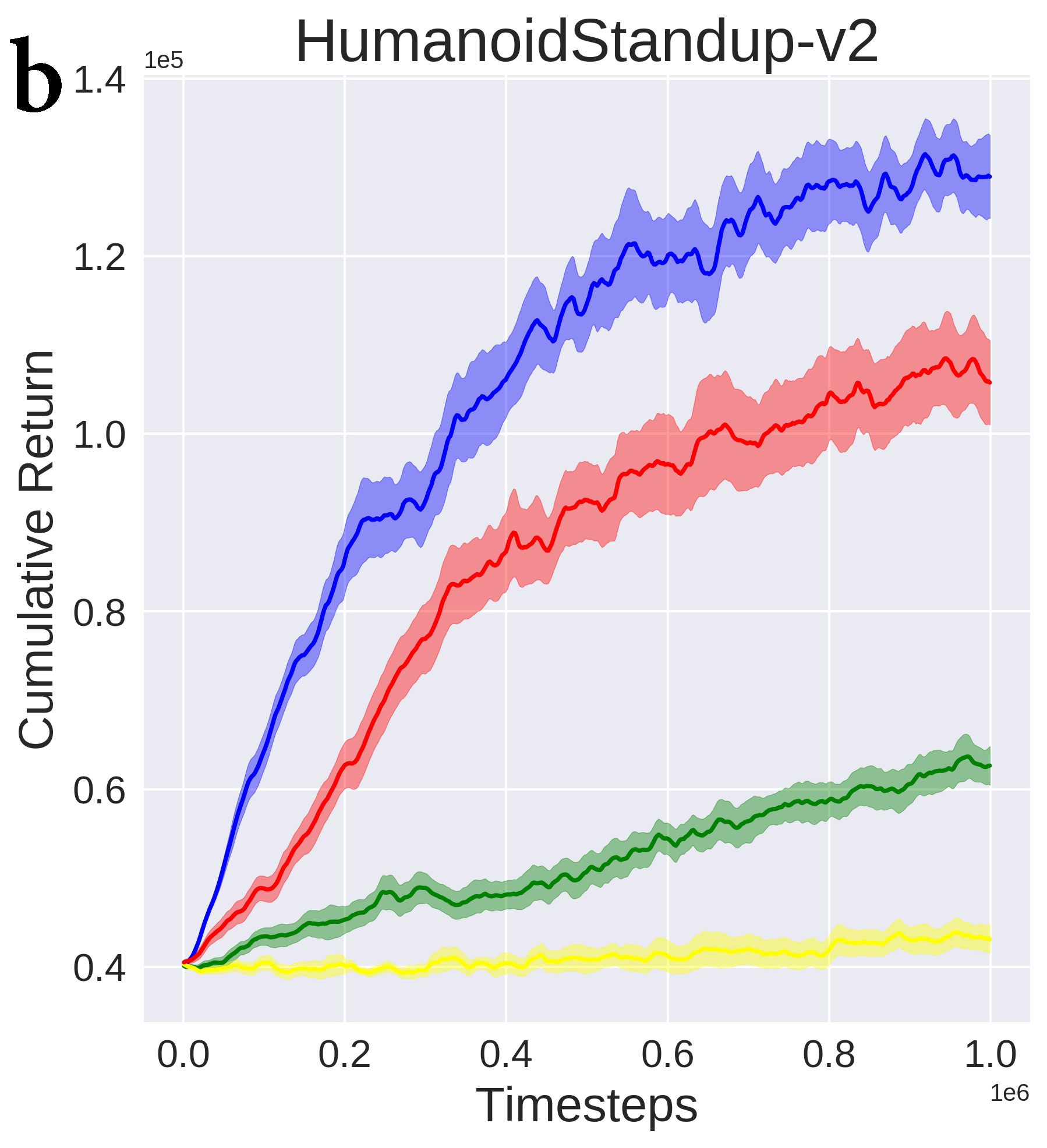}
  \end{subfigure}
  \begin{subfigure}{.33\textwidth}
  \centering
  \includegraphics[width=.8\linewidth]{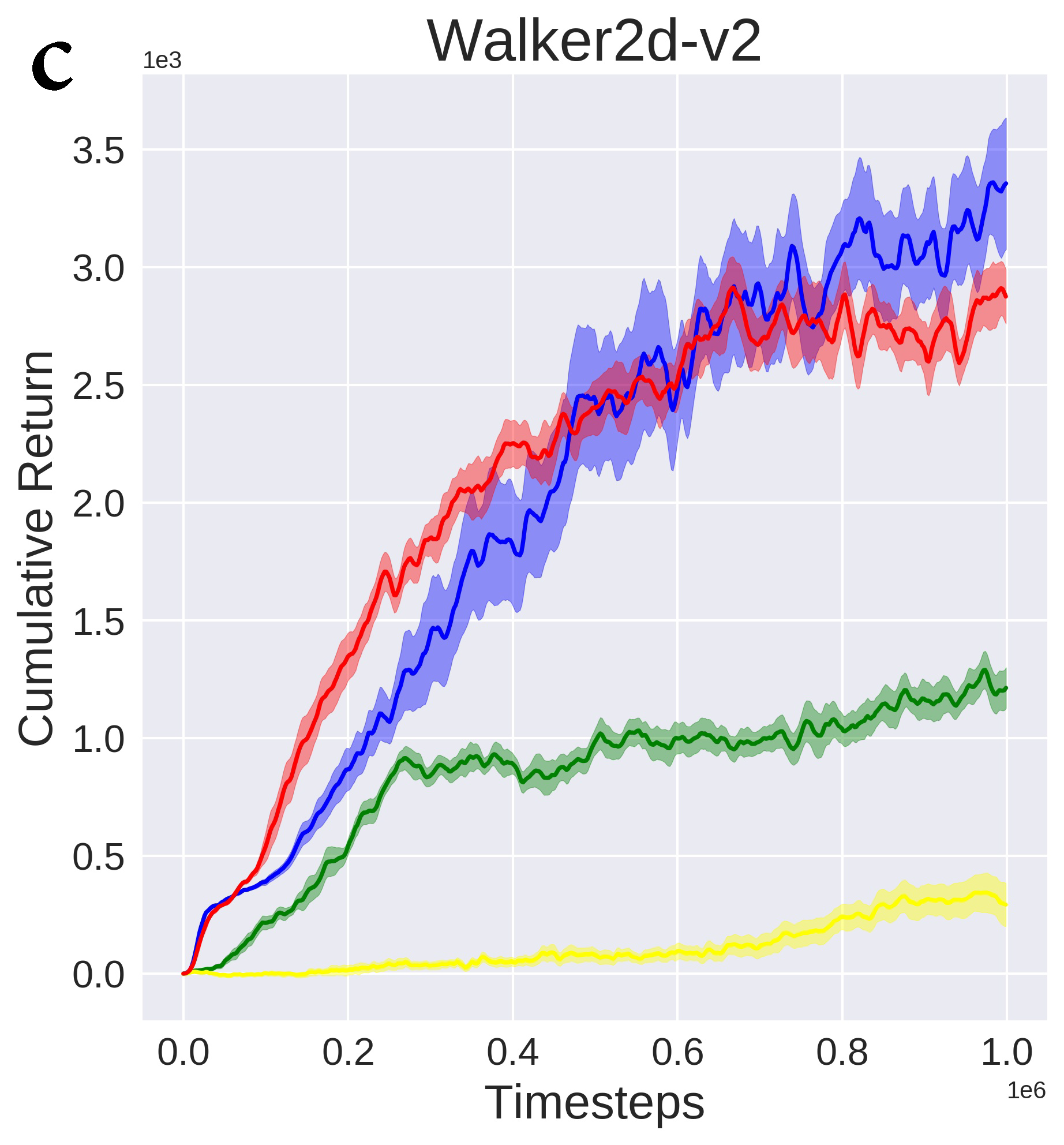}
  \end{subfigure}
  \caption{Comparison of different scales of temperature parameter $\eta$ on the performance of SPOD in three high-dimensional control tasks. The greater $\eta$ indicates the agent is likely to explore the new more returnable policies in the early training stage, and vice versa.}\label{comp_ent}
\end{figure*}

\textbf{Temperature parameter:} From Eq.~\ref{entropy goal}, temperature parameter $\eta$ determines the relative importance of reward and entropy. Thus, it balances the relationship of exploration and exploitation. Through setting $\eta$ as linear decay, the agent will tend to explore the new policy in the early training stage and exploit the explored policy in the final training stage. Fig.~\ref{comp_ent} shows the performance of SPOD under variable $\eta$. If $\eta$ is too large, the agent is so addicted to explore the new policy that ignores exploiting the reward signal, and consequently fails to improve its performance. Conversely, if $\eta$ is too small, SPOD will degenerate to PPO and the policy will quickly becomes deterministic. Although we have achieved the outstanding results by using the fixed temperature parameter $\eta = 1e^{-3}$ in Mujoco, it is recommended to determine the value scale of $\eta$ based on the ratio of reward and entropy in practical environments.

\subsection{Ablation studies}
\begin{figure}
  \centering
  \includegraphics[width=5cm]{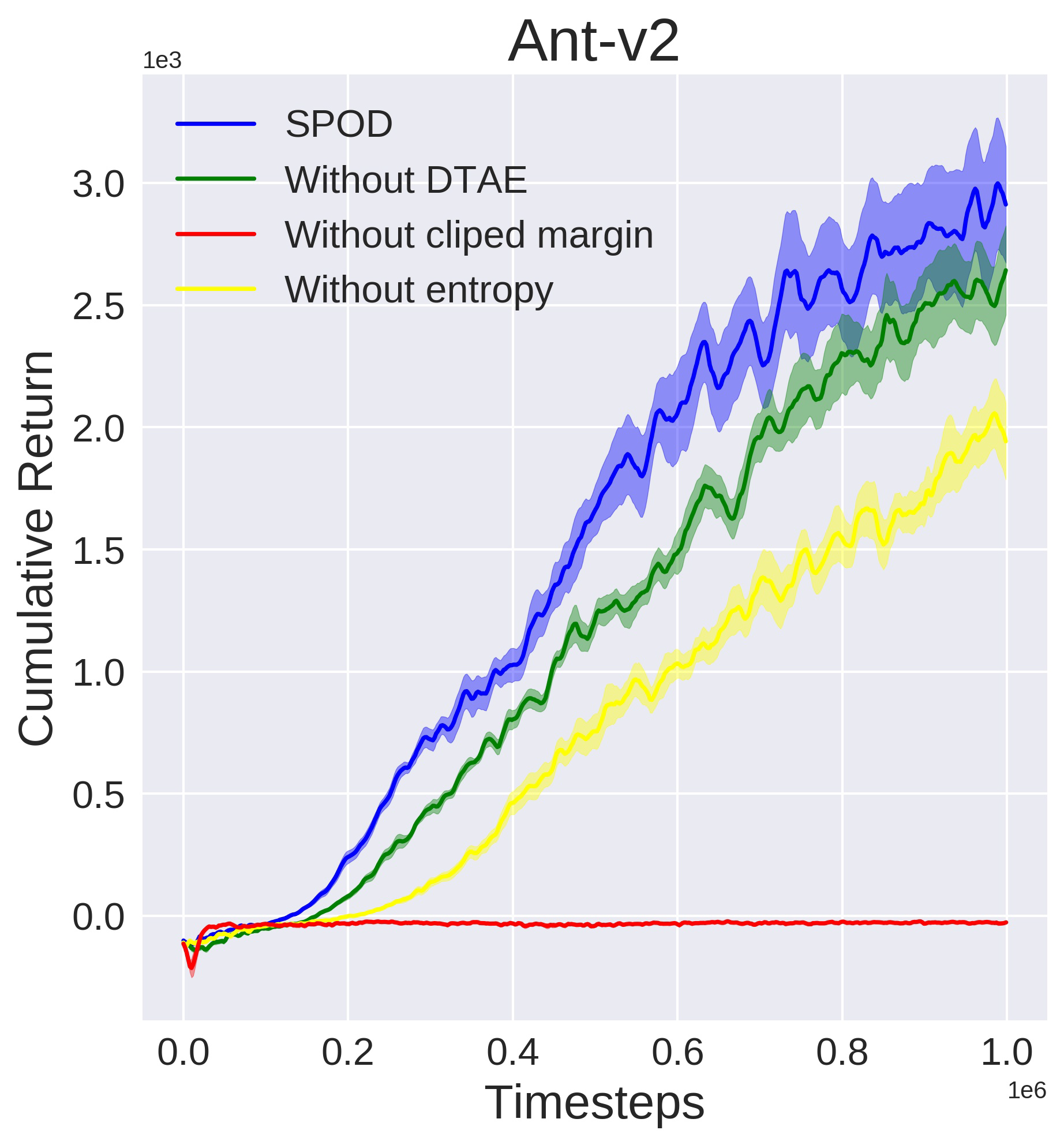}
  \caption{Ablation analysis of SPOD}\label{comp_abl}
\end{figure}
In this experiment, using the control variable method, we ablate three core components of SPOD: dual-track advantage estimator (DTAE), entropy term and clipped margin respectively, to quantify their contribution to the overall performance of SPOD. First, without DTAE, the agent will adopt GAE to estimate the advantage of one action at state $s$. Fig.~\ref{comp_abl} shows DTAE can acquire high cumulative return as well as keep faster training speed compared with GAE. Then, without entropy term in Eq.~\ref{cobjective}, SPOD will degenerate to PPO and the algorithm's performance will also decrease. Finally, without the clipped margin, cumulative return remains extremely low level throughout the training process and the policy is rarely optimized since the prerequisite for the algorithm to converge is that the new explored policy and the old policy should locate at the clipped margin $\epsilon$ or trust region $\delta$ (Appendix~\ref{converge}).

\section{Conclusion}
Since entropy can balance the opportunity of exploration and exploitation in reinforcement learning, in this paper, we have introduced it to the objective function to optimize policy. Theoretically, we have proved that the proposed algorithm can improve the policy in each iteration. Experimentally, we have illustrated that the agent controlled by our algorithm performs more excellent than the classical on-policy algorithms in benchmark environments. Nevertheless, besides improving policy, entropy will also cause the low training speed because the agent tends to adopt the exploratory action rather than the greedy action in the initial phase. In this case, we proposed the dual-track advantage estimator to accelerate the convergence of the entropy-based optimization methods. Results show the integrating of DTAE and entropy can obviously increase cumulative return and training speed.

\section{Appendix}
\subsection{The difference in policy performance}\label{difference}
Given two policy $\pi$ and $\hat{\pi}$. We have $A^H_{\pi}(s_t,a_t) =  \mathbb{E}_{s_{t+1} \sim P(s_{t+1}|s_t,a_t)}[r^H(s_t,a_t) + \gamma V_{\pi}^H(s_{t+1}) - V_{\pi}^H(s_t)]$. Then,
\begin{align*}
    & \mathbb{E}_{\tau | \hat{\pi}} [\sum_{t=0}^{\infty} \gamma^t A^H_{\pi}(s_t,a_t)] \\
    & = \mathbb{E}_{\tau | \hat{\pi}} [\sum_{t=0}^{\infty} \gamma^t(r^H_{\pi}(s_t.a_t) + \gamma V^H_{\pi}(s_{t+1}) - V^H_{\pi}(s_t))] \\
    & = \mathbb{E}_{\tau | \hat{\pi}} [\sum_{t=0}^{\infty} \gamma^t(r(s_t.a_t) + \eta H_{\pi}(s_{t+1})+ \gamma V^H_{\pi}(s_{t+1}) - V^H_{\pi}(s_t))] \\
    & = -\mathbb{E}_{s_0} [V_{\pi}^H(s_0)] + \mathbb{E}_{\tau | \hat{\pi}} [\sum_{t=0}^{\infty} \gamma^t(r(s_t.a_t) + \eta H_{\pi}(s_{t+1}) \\
    & \quad - \eta H_{\hat{\pi}}(s_{t+1}) + \eta H_{\pi}(s_{t+1}))] \\
    & = -\mathbb{E}_{s_0} [V_{\pi}^H(s_0)] + \mathbb{E}_{\tau | \hat{\pi}}[\sum_{t=0}^{\infty} \gamma^t (r^H_{\hat{\pi}}(s_t,a_t)+ \eta (H_{\pi}(s_{t+1})  \\
    & \quad - H_{\hat{\pi}}(s_{t+1})))] \\
    & = -J(\pi) + J(\hat{\pi}) + \mathbb{E}_{\tau | \hat{\pi}}[\sum_{t=0}^{\infty} \eta \gamma^t(H_{\pi}(s_{t+1}) - H_{\hat{\pi}}(s_{t+1}))]
\end{align*}
Define $T_{\pi}(s_t,a_t) = A^H_{\pi}(s_t, a_t) + \eta [H_{\hat{\pi}}(s_{t+1}) - H_{\pi}(s_{t+1})]$, we have:
\begin{equation*}
J(\hat{\pi}) - J(\pi) = \mathbb{E}_{\tau|\hat{\pi}} \gamma^t T_{\pi} (s_t,a_t)
\end{equation*}

\subsection{Proof of policy performance bound}\label{pbound}
Consider $\kappa$-coupled policies $(\pi, \hat{\pi})$, we have:
\begin{equation}\label{JL}
  \begin{split}
       & J(\hat{\pi}) = J(\pi) + \mathbb{E}_{\tau|\hat{\pi}} \gamma^t \hat{T}(s_t) \\
       & L_{\pi}(\hat{\pi}) = J(\pi) + \mathbb{E}_{\tau|\pi} \gamma^t \hat{T}(s_t)
  \end{split}
\end{equation}
In this paper, we define $\kappa = D_{TV}^{max}(\pi, \hat{\pi}) = \max_s \frac{1}{2} \sum_{a}|\pi(a|s) - \hat{\pi}(a|s)|$ and the trust region $\delta \geq D_{KL}(\pi || \hat{\pi})$. Before derive the bound of $J(\hat{\pi})$, we firstly present three lemmas:

\begin{lemma} \label{entropybound}
  Consider two policies $(\pi, \hat{\pi})$ located in the trust region, that is $D_{KL}(\pi || \hat{\pi}) \leq \delta$, where $\pi$ and $\hat{\pi}$ are two normal distribution: $\pi = \mathcal{N}(\mu, \sigma), \hat{\pi} = \mathcal{N}(\hat{\mu}, \hat{\sigma})$. The bound of their entropy difference holds: $H_{\hat{\pi}}(s) - H_{\pi}(s) \leq \delta + \frac{1}{2}\log e$.
  \begin{align*}
   & D_{KL} (\pi || \hat{\pi}) = \int_{a} \pi(a|s) \log \frac{\pi(a|s)}{\hat{\pi}(a|s)} \\
   & = \int_a \pi(a|s) \left[\log \frac{\hat{\sigma}}{\sigma} + (\frac{(a-\hat{\mu})^2}{2\hat{\sigma}^2} - \frac{(a-\mu)^2}{2\sigma^2})\log e \right ] da \\
   & = \log \frac{\hat{\sigma}}{\sigma} + \log e \int_a \pi(a|s) \left(\frac{(a-\hat{\mu})^2}{2\hat{\sigma}^2} - \frac{(a-\mu)^2}{2\sigma^2} \right) da \\
   & = \log \frac{\hat{\sigma}}{\sigma} + \log e \int_a \pi(a|s) \frac{(a-\hat{\mu})^2}{2\hat{\sigma}^2} da \\
   & - \log e \int_a \pi(a|s) \frac{(a-\mu)^2}{2\sigma^2} da  \\
   & = \log \frac{\hat{\sigma}}{\sigma} + \log e \int_a \pi(a|s) \frac{(a-\hat{\mu})^2}{2\hat{\sigma}^2} da - \frac{1}{2}\log e \leq \delta  \\
   & \Rightarrow \log \frac{\hat{\sigma}}{\sigma} - \frac{1}{2}\log e \leq \delta \\
   & \Rightarrow \log \frac{\hat{\sigma}}{\sigma} \leq \delta + \frac{1}{2}\log e \\
   & H_{\pi}(s) = - \int_a \pi(a|s) \log \pi(a|s) da\\
   & = - \int_a \pi(a|s) \log \frac{1}{\sqrt{2 \pi \sigma^2}} e^{-\frac{(a-\mu)^2}{2\sigma^2}} da\\
   & = -\int_a \pi(a|s)\log \frac{1}{\sqrt{2 \pi \sigma^2}} - \log e \int_a \pi(a|s) \frac{(a-\mu)^2}{2\sigma^2} da \\
   & = \frac{1}{2} \log 2\pi\sigma^2 + \log e \times \frac{\sigma^2}{2\sigma^2} \\
   & =\frac{1}{2} \log (2\pi e \sigma^2) \\
   & H_{\hat{\pi}}(s) -H_{\pi}(s)= \frac{1}{2} \log (2\pi e \hat{\sigma}^2) - \frac{1}{2} \log (2\pi e \sigma^2) \\
   & = \log \frac{\hat{\sigma}}{\sigma} \leq  \delta + \frac{1}{2}\log e
  \end{align*}
\end{lemma}
\begin{lemma} \label{Tbound}
Given that $(\pi, \hat{\pi})$ are $\kappa$-coupled policies located in the trust region, we define $\xi = 2 \max_{s,a}|A^H_{\pi}(s,a)| + \frac{\eta}{\kappa}(\delta + \frac{1}{2}\log e)$, for all $s$,
\begin{equation}\label{upbound}
|\hat{T}(s)| \leq \kappa\xi
\end{equation}
\begin{proof}
\begin{align*}
    & \mathbb{E}_{a \sim \pi} A^H_{\pi}(s,a) = 0 \\
    &\hat{T}(s) = \mathbb{E}_{\hat{a}\sim\hat{\pi}} [T_{\pi}(s, \hat{a})] = \mathbb{E}_{(a,\hat{a})\sim(\pi, \hat{\pi})} [T_{\pi}(s,\hat{a}) - A^H_{\pi}(s,a)] \\
    & =\mathbb{E}_{(a,\hat{a})\sim(\pi, \hat{\pi})} [A^H_{\pi}(s,\hat{a}) - A^H_{\pi}(s,a) + \eta \mathbb{E}_{\hat{a}} (H_{\hat{\pi}}(s')- H_{\pi}(s'))] \\
    & = p(a \neq \hat{a}) \mathbb{E}_{(a,\hat{a}\sim(\pi, \hat{\pi}))} [A^H_{\pi}(s,\hat{a}) - A^H_{\pi}(s,a)] \\
    & + \eta \mathbb{E}_{\hat{a}} (H_{\hat{\pi}}(s')- H_{\pi}(s')) \\
    & \leq \kappa \cdot 2 \max_{s,a}|A^H_{\pi}(s,a)| + \eta(\delta + \frac{1}{2}\log e) = \kappa\xi \quad (Lemma~\ref{entropybound})
\end{align*}
\end{proof}
\end{lemma}

\begin{lemma}\label{couplebound}
$(\pi, \hat{\pi})$ are two $\kappa$-coupled policies located in trust region, then:
\begin{equation*}
  |\mathbb{E}_{s_t \sim \hat{\pi}}\hat{T}(s_t) - \mathbb{E}_{s_t \sim \pi} \hat{T}(s_t)| \leq 2(1-(1-\kappa)^t) \xi
\end{equation*}
\begin{proof}
Under the same time seed, we generate two trajectories $\tau$ and $\hat{\tau}$ based on $\pi$ and $\hat{\pi}$ respectively. Note that $\pi$ and $\hat{\pi} $ are $\kappa$-coupled policies. It means the trajectories $\tau$ and $\hat{\tau}$ are very consistent and the probability that actions $a_t$ is not agree with $\hat{a}_t$ at time $t$ is $p(a_t \neq \hat{a}_t) \leq \kappa$. Let $n_t$ denotes the times that two actions are inconsistent ($a_i \neq \hat{a}_i, i \leq t$) in the two trajectories before time $t$, we have:
\begin{align*}
    \mathbb{E}_{s_t \sim \hat{\pi}}[\hat{T}(s_t)] &   = p(n_t = 0) \mathbb{E}_{s_t \sim \hat{\pi}|n_t=0}[\hat{T}(s_t)] \\
     & + p(n_t>0) \mathbb{E}_{s_t \sim \hat{\pi}|n_t>0}[\hat{T}(s_t)] \\
    \mathbb{E}_{s_t \sim \pi}[\hat{T}(s_t)] &  = p(n_t = 0) \mathbb{E}_{s_t \sim \pi|n_t=0}[\hat{T}(s_t)] \\
     & + p(n_t>0) \mathbb{E}_{s_t \sim \pi|n_t>0}[\hat{T}(s_t)]
\end{align*}
$n_t=0$ means the trajectories $\tau$ and $\hat{\tau}$ are completely coincident before time $t$, then:
\begin{equation*}
  p(n_t = 0) \mathbb{E}_{s_t \sim \hat{\pi}|n_t=0}[\hat{T}(s_t)] = p(n_t = 0) \mathbb{E}_{s_t \sim \pi|n_t=0}[\hat{T}(s_t)]
\end{equation*}
At this time, we have:
\begin{align*}
  & \mathbb{E}_{s_t \sim \hat{\pi}}[\hat{T}(s_t)] - \mathbb{E}_{s_t \sim \pi}[\hat{T}(s_t)] \\
  & = p(n_t>0) \left( \mathbb{E}_{s_t \sim \hat{\pi}|n_t>0}[\hat{T}(s_t)] - \mathbb{E}_{s_t \sim \pi|n_t>0}[\hat{T}(s_t)] \right)
\end{align*}
Duo to $(\pi, \hat{\pi})$ are $\kappa$-coupled policies, $p(a_t \neq \hat{a}_t) \leq \kappa$, and then $p(a_t = \hat{a}_t) \geq 1-\kappa$. In RL, it is reasonable to assume that sampling the action $a$ from the policy $\pi$ is an independent event at each time using Monte Carlo method. Therefore, $p(n_t = 0) = p(a_1 = \hat{a}_1)p(a_2 = \hat{a}_2)\cdot\cdot\cdotp(a_t = \hat{a}_t) \geq (1-\kappa)^t$. And its opposite event $p(n_t \neq 0) \leq 1 -(1-\kappa)^t$. We can derive:
\begin{align*}
& |\mathbb{E}_{s_t \sim \hat{\pi}}[\hat{T}(s_t)] - \mathbb{E}_{s_t \sim \pi}[\hat{T}(s_t)]| \\
& \leq (1 -(1-\kappa)^t) |\mathbb{E}_{s_t \sim \hat{\pi}|n_t>0}[\hat{T}(s_t)] - \mathbb{E}_{s_t \sim \pi|n_t>0}[\hat{T}(s_t)]| \\
& \leq (1 -(1-\kappa)^t)\{|\mathbb{E}_{s_t \sim \hat{\pi}|n_t>0}[\hat{T}(s_t)]| + |\mathbb{E}_{s_t \sim \pi|n_t>0}[\hat{T}(s_t)]|\}\\
& \leq 2(1 -(1-\kappa)^t) \max_{s}|\hat{T}(s)| \\
& \leq 2(1 -(1-\kappa)^t) \kappa\xi \quad (Lemma~\ref{Tbound})
\end{align*}
\end{proof}
\end{lemma}
According to the above three Lemmas, we can demonstrate the performance bound of policy $\hat{\pi}$. Based on Eq.~\ref{JL}, we have:
\begin{align*}
       & |J(\hat{\pi}) - L_{\pi}(\hat{\pi})| = \sum_{t=0}^{\infty}\gamma^t|\mathbb{E}_{\tau|\hat{\pi}} \gamma^t \hat{T}(s_t) - \mathbb{E}_{\tau|\pi} \gamma^t \hat{T}(s_t)| \\
       & \leq \sum_{t=0}^{\infty} \gamma^t \cdot 2(1 -(1-\kappa)^t) \kappa\xi = \frac{2\kappa\xi}{1-\gamma} - \frac{2\kappa\xi}{1- \gamma(1-\kappa)} \\
       & = \frac{2\xi\gamma\kappa^2}{(1-\gamma)(1-\gamma(1-\kappa))} \leq \frac{2\xi\gamma\kappa^2}{(1-\gamma)^2}
\end{align*}
\cite{pollard2000asymptopia} proved $D_{TV}(\pi||\hat{\pi})^2 = \kappa^2 \leq D_{KL}(\pi||\hat{\pi})$. Define $D_{KL}^{\max}(\pi,\hat{\pi}) = \max_s D_{KL}(\pi(\cdot|s)||\hat{\pi}(\cdot|s))$, we have:
\begin{align*}
       & |J(\hat{\pi}) - L_{\pi}(\hat{\pi})| \leq \frac{2\xi\gamma}{(1-\gamma)^2} D_{KL}^{\max}(\pi,\hat{\pi}) \\
       & J(\hat{\pi}) \geq L_{\pi}(\hat{\pi}) - \frac{2\xi\gamma}{(1-\gamma)^2} D_{KL}^{\max}(\pi,\hat{\pi})
\end{align*}

\subsection{Trust region policy optimization}\label{converge}
From the above derivation, our objective is :
\begin{equation*}
 \max_{\hat{\pi}} \left[ L_{\pi_{old}}(\hat{\pi}) - C D_{KL}^{\max}(\pi_{old}, \hat{\pi}) \right]
\end{equation*}
Duo to the coupled policies $(\pi, \hat{\pi})$ are located in the trust region ($D_{KL}(\pi || \hat{\pi}) \leq \delta$), we can simplify the objective function as:
\begin{align*}
       & \max_{\hat{\pi}} \quad L_{\pi_{old}}(\hat{\pi}) = J(\pi_{old}) + \mathbb{E}_{\tau|\pi_{old}} \gamma^t \hat{T}(s_t)\\
       & \qquad s.t. \quad D_{KL}(\pi_{old} || \hat{\pi}) \leq \delta
\end{align*}
where $J(\pi_{old})$ is a constant. Define the discounted visitation frequencies:
\begin{equation*}
  \rho_{\pi}(s) = P_{\pi}(s_0=s) + \gamma P_{\pi}(s_1=s) + \gamma^2 P_{\pi}(s_2=s) + \cdot \cdot \cdot
\end{equation*}
\begin{align*}
   \sum_s \rho_{\pi}(s) & = \sum_s P_{\pi}(s_0=s) + \gamma \sum_s P_{\pi}(s_1=s) + \cdot \cdot \cdot \\
   & = 1 + \gamma + \gamma^2 + \cdot \cdot \cdot = \frac{1}{1-\gamma}
\end{align*}
so the objective function equates to:
\begin{align*}
    \max_{\hat{\pi}} & \quad \mathbb{E}_{\tau|\pi_{old}} \gamma^t \hat{T}(s_t) = \mathbb{E}_{\tau|\pi_{old}} \gamma^t \mathbb{E}_{a_t\sim \hat{\pi}(\cdot|s_t)} T_{\pi_{old}} (s_t,a_t)\\
    & = \mathbb{E}_{\tau|\pi_{old}} \gamma^t \sum_{a_t} \pi_{old}(a_t|s_t) \frac{\hat{\pi}(a_t|s_t)}{\pi_{old}(a_t|s_t)} T_{\pi_{old}} (s_t,a_t) \\
    & = \mathbb{E}_{\tau|\pi_{old}} \gamma^t \mathbb{E}_{a_t \sim \pi_{old}} \frac{\hat{\pi}(a_t|s_t)}{\pi_{old}(a_t|s_t)} T_{\pi_{old}} (s_t,a_t) \\
    & = \sum_{t=0}^{\infty} \sum_{s}\gamma^t P_{\pi_{old}}(s_t =s) \mathbb{E}_{a\sim \pi_{old}} \frac{\hat{\pi}(a|s)}{\pi_{old}(a|s)} T_{\pi_{old}} (s,a) \\
    & = \sum_{s}\sum_{t=0}^{\infty}\gamma^t P_{\pi_{old}}(s_t =s) \mathbb{E}_{a\sim \pi_{old}} \frac{\hat{\pi}(a|s)}{\pi_{old}(a|s)} T_{\pi_{old}} (s,a) \\
    & = \sum_{s}\rho_{\pi_{old}}(s) \mathbb{E}_{a\sim \pi_{old}} \frac{\hat{\pi}(a|s)}{\pi_{old}(a|s)} T_{\pi_{old}} (s,a)  \\
   \Upsilon \rightarrow & = (1-\gamma)\sum_{s} \frac{\rho_{\pi_{old}}(s)}{1-\gamma} \mathbb{E}_{a\sim \pi_{old}} \frac{\hat{\pi}(a|s)}{\pi_{old}(a|s)} T_{\pi_{old}} (s,a) \\
    & = (1-\gamma) \mathbb{E}_{s \sim \rho_{old}, a \sim \pi_{old}} \frac{\hat{\pi}(a|s)}{\pi_{old}(a|s)} T_{\pi_{old}} (s,a) \\
    & \qquad s.t. \quad D_{KL}(\pi_{old} || \hat{\pi}) \leq \delta
\end{align*}
In practice, constant $(1-\gamma)$ has no influence on the optimization results and $s,a$ is generated by Monte Carlo Method. Therefore, the optimization objective can be transformed into:
\begin{align*}
    \max_{\hat{\theta}} \quad & \hat{\mathbb{E}}_t \left[ \frac{\pi_{\hat{\theta}}(a_t|s_t)}{\pi_{old}(a_t|s_t)} T_{\pi_{old}}(s_t, a_t) \right] \\
    & s.t. \quad \hat{\mathbb{E}}_t [D_{KL}(\pi_{old}(\cdot|s) || \hat{\pi}(\cdot|s))] \leq \delta
\end{align*}
Note that although the final objective contains $t$, the samples $(s_t,a_t,r_{t+1}, s_{t+1})$ generated by policy $\pi$ are time-independent in the optimization process since we have eliminatd the time series of samples in the proof step $\Upsilon$ and $t$ is just used for labelling the samples.
\subsection{Supplement for reproducibility}\label{supply}
To reproduce the results of this article, we show the implementation details of the proposed algorithm (SPOD) as follows.
\subsubsection*{Software version}
During the algorithm implementation, we encountered many software adaptation problems. For the sake of convenience, Table~\ref{software} lists the softwares and their version we used in our experiments.

\begin{table}[H]
\center
\caption{Softwares}
\begin{tabular}{l|c}
  \hline
  Software & Version \\
  \hline
  Ubuntu & 18.04 \\
  Python & 3.6.8 \\
  Tensorflow & 1.8.0 \\
  Mujoco & mjpro 150 linux \\
  Mujoco-py & 1.50\\
  Gym & 1.1.4 \\
  \hline
\end{tabular}\label{software}
\end{table}

\subsubsection*{Pseudo-code and open-source code}
Algorithm~\ref{algo} shows the pseudo-code of SPOD. Additionally, we have released the corresponding Python code on GitHub\footnote{https://github.com/Code-Papers/SPOD}.
%The results of A2C, TRPO, PPO in Fig.~\ref{comp_methods} are plotted by the OpenAI baseline repository\footnote{https://github.com/openai/baselines}.
\begin{algorithm}
\caption{SPOD: soft policy optimization using DTAE}\label{SPOD}
\textbf{Input:} initialize policy parameters $\theta_0$, shadow policy parameters $\tilde{\theta}_0$, value function parameters $\phi_0$ and shadow value function $\tilde{\phi}_0$.

 \begin{algorithmic}[1]
   \FOR {$k = 0$ to $K$}
   \STATE Collect set of trajectories $D_{k} = \{\tau_{i}\}$ by running policy $\pi_{k} = \pi(\theta_{k})$ in the environment.
   \STATE Compute rewards-to-go $G_t$ based on $D_k$
   \STATE Compute the TD errors, $\delta^{\theta_k}_t$ and $\delta^{\tilde{\theta}_k}_t$ based on $V_{\phi_k}$, $V_{\tilde{\phi}_{k}}$ and $\pi_{\theta_k}$, $\pi_{\tilde{\theta}_k}$ respectively (Eq.~\ref{sTD}). And then compute the entropy-based advantage function $T_t$ according to Eq.~\ref{T}.
   \STATE $\tilde{\theta}_{k+1}  = \theta_k$, $\tilde{\phi}_{k+1} = \phi_k$
   \STATE Update the policy by maximizing the PPO-Clip objective:
          \begin{equation*}
          \begin{split}
              & \theta_{k+1} = \arg \max_{\theta} \frac{1}{|D_{k}|} \sum_{\tau \in D_k} \sum_{t = 0} \\
              & \min(\frac{\pi_\theta(a_t|s_t)}{\pi_{\theta_k}(a_t|s_t)} T_t, g(\epsilon, T_t))
          \end{split}
          \end{equation*}
   \STATE Fit value function by regression on mean-squared error:
          \begin{equation*}
            \phi_{k+1} = \arg \min_{\phi} \frac{1}{|D_{k}|T} \sum_{\tau \in D_k} \sum_{t = 0}^{T}(G^H_t - V^H_{\phi_{k}}(s_t))^2
          \end{equation*}
   \ENDFOR
 \end{algorithmic}\label{algo}
\end{algorithm}

\subsubsection*{Hyper-parameters}
Table~\ref{hyper} lists the default hyper-parameters we used in the experiments.
\begin{table}[H]
\center
\caption{The hyper-parameters used in SPOD}
\begin{tabular}{ll}
\hline
\multicolumn{2}{c}{Common Hyper-parameters}                                     \\ \hline
\multicolumn{1}{l|}{Neural network} & MLP \\
\multicolumn{1}{l|}{Activation function} & ReLU \\
\multicolumn{1}{l|}{Learning rate} & $3e^{-4}$ (Linear decay) \\
\multicolumn{1}{l|}{Discount($\gamma$)}      & 0.99               \\
\multicolumn{1}{l|}{GAE($\lambda$)}          & 0.95               \\
\multicolumn{1}{l|}{Hidden layer number}                  & 2                  \\
\multicolumn{1}{l|}{Hidden units per layer}               & 64                 \\
\multicolumn{1}{l|}{Minibatch size}                       & 64                 \\
\multicolumn{1}{l|}{Optimizer}                            & Adam               \\
\multicolumn{1}{l|}{Value loss coefficient}               & 0.5                \\ \hline
\multicolumn{2}{c}{SPOD Hyperparameters}                                      \\ \hline
\multicolumn{1}{l|}{DTAE combination type}              & mean                \\
\multicolumn{1}{l|}{Entropy loss coefficient}             & 1                  \\
\multicolumn{1}{l|}{Clipped margin $\epsilon$}            & 0.2 (Linear decay)    \\
\multicolumn{1}{l|}{Temperature parameter $\eta$}             & $1e^{-3}$ (Linear decay) \\
\multicolumn{1}{l|}{TD update coefficient $\alpha$}              & 0.1                \\
\hline
\end{tabular}\label{hyper}
\end{table}

%\section*{Acknowledgment}
%
%The preferred spelling of the word ``acknowledgment'' in America is without
%an ``e'' after the ``g''. Avoid the stilted expression ``one of us (R. B.
%G.) thanks $\ldots$''. Instead, try ``R. B. G. thanks$\ldots$''. Put sponsor
%acknowledgments in the unnumbered footnote on the first page.

\bibliographystyle{IEEEtran}
% argument is your BibTeX string definitions and bibliography database(s)
\bibliography{IEEEexample}

\end{document}